\documentclass[]{amap}

\usepackage[toc,page,header]{appendix}
\usepackage{minitoc}
\usepackage{solarized-light}
\usepackage{booktabs}
\usepackage{multirow}
\usepackage{graphicx}
\usepackage{array}
\usepackage{siunitx,array}
\usepackage[table]{xcolor}
\usepackage{makecell,array,booktabs}
\usepackage{makecell}
\usepackage{framed}

\usepackage{bbding}
\usepackage{hyperref}
\usepackage{amsmath}
\usepackage{amsfonts}
\usepackage{placeins}
\usepackage[colorinlistoftodos]{todonotes}
\usepackage{longtable}
\usepackage{hhline}
\usepackage{fancyvrb}
\usepackage{graphicx}
\usepackage[table]{xcolor}
\usepackage{booktabs}
\usepackage{float}
\usepackage{fvextra}
\usepackage{CJKutf8}
\usepackage{multicol}
\usepackage{float}
\usepackage{placeins}
\usepackage{cleveref}
\usepackage{tablefootnote}
\usepackage{threeparttable}
\usepackage{tabularx}
\usepackage{mdframed}
\usepackage{subcaption}
\usepackage[usestackEOL]{stackengine}
\usepackage[numbers]{natbib}
\newcommand{\commentout}[1]{}
\renewcommand{\paragraph}[1]{\noindent\textbf{#1.}\hspace*{1em}}
\usepackage{enumitem}
\usepackage{hyperref}
\setlist[itemize]{leftmargin=15pt}
\usepackage{siunitx,array}
\usepackage{tabularx}
\usepackage{adjustbox}
\usepackage{arydshln}
\usepackage{pifont}
\usepackage{bbding}
\definecolor{ampblue}{rgb}{0.82, 0.88, 0.94}
\usepackage[table]{xcolor} 
\usepackage{array} 
\usepackage[dvipsnames]{xcolor}

\RequirePackage{xspace}
\makeatletter
\DeclareRobustCommand\onedot{\futurelet\@let@token\@onedot}
\def\@onedot{\ifx\@let@token.\else.\null\fi\xspace}

\makeatother
\newcommand{\ABotWAM}{ABot-M0.5}

\usepackage{xcolor}

\definecolor{abot1}{HTML}{0185FE}
\definecolor{abot2}{HTML}{0185FE}
\definecolor{abot3}{HTML}{0185FE}
\definecolor{abot4}{HTML}{0185FE}
\definecolor{abot5}{HTML}{FB8C00}
\definecolor{abot6}{HTML}{FB8C00}
\definecolor{abot7}{HTML}{FB8C00}

\newcommand{\ABotMZeroFive}{%
{\color{abot1}A}%
{\color{abot2}B}%
{\color{abot3}o}%
{\color{abot4}t}%
{\color{abot5}-}%
{\color{abot6}M}%
{\color{abot7}0.5}%
}

\title{\ABotMZeroFive: Unified Mobility-and-Manipulation World Action Model}

\author{AMAP CV Lab}
\vspace{-11pt}

\contribution{See \hyperref[sec:contributions]{Contributions} section for a full author list.}

\abstract{

Mobile manipulation is a key capability for general-purpose robots, yet remains challenging for current embodied learning methods. VLA policies are typically reactive and lack explicit world modeling, while existing World Action Models (WAMs) are still poorly aligned with the structure of mobile manipulation: they operate on coarse video chunks, model entangled navigation-manipulation actions, and train inverse dynamics under supervision that does not match autoregressive inference. As a result, they often miss fine-grained contact dynamics, suffer from action-distribution conflicts, and accumulate errors over long-horizon rollouts.

We propose \textbf{ABot-M0.5}, a new WAM built on the insight that \textit{mobile manipulation requires alignment at three levels: temporal granularity, action space, and train-test consistency.} To align temporal granularity, we introduce intermediate latent actions that capture local visual state transitions and serve as an bridging action space between video latents and embodiment-specific controls. To align action space, we design a dual-level Mixture-of-Transformers architecture that disentangles both modality representations and heterogeneous action subspaces such as base movement and arm manipulation. To align inference conditions, we propose the dream-forcing training strategy that progressively trains inverse dynamics on model-predicted videos, improving train-test alignment and robustness during autoregressive prediction.

Experiments on challenging mobile and fine-grained manipulation benchmarks demonstrate that ABot-M0.5 achieves state-of-the-art performance in both long-horizon task success and fine-grained control accuracy. These results highlight the critical importance of granularity-aligned, action-disentangled, and inference-consistent world-action modeling.

\bigskip

\textbf{Date:} 
July 1, 2026

\textbf{Code:} 
\url{https://github.com/amap-cvlab/ABot-Manipulation}

}

\begin{document}
\maketitle
\vspace{-4pt}

\begin{figure}[!h]
    \centering
    \vspace{-10pt}
\includegraphics[width=0.6\linewidth]{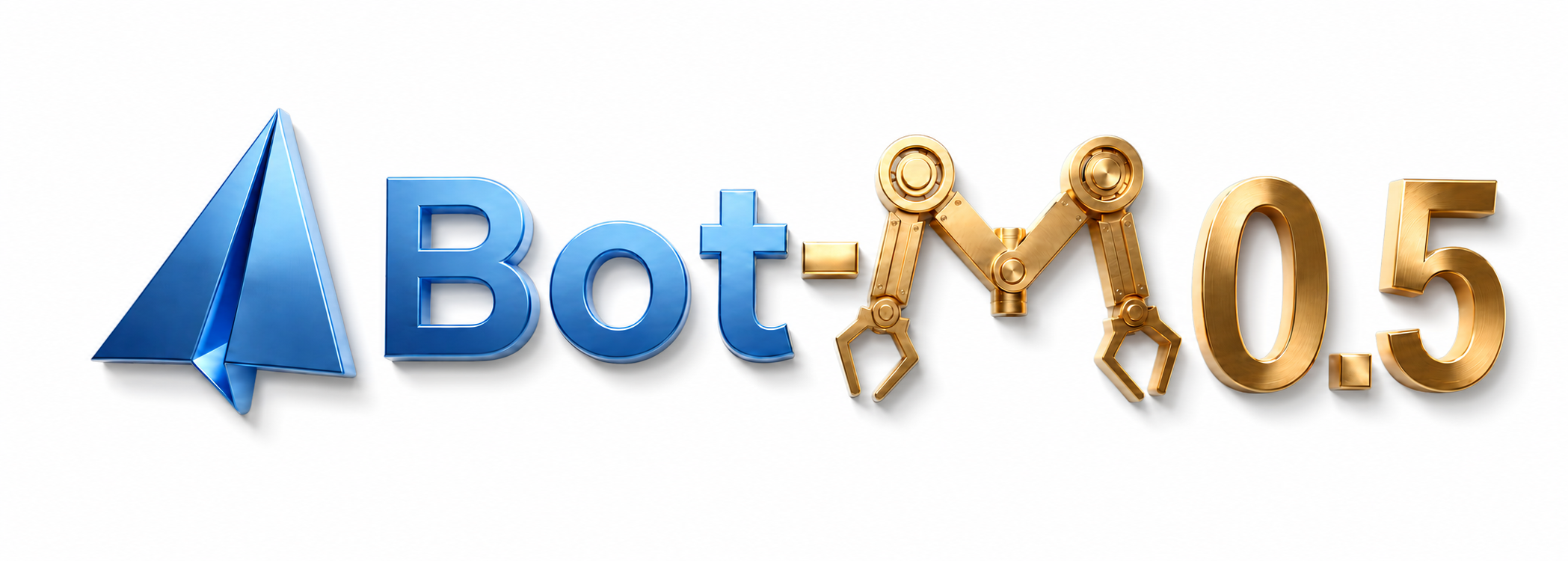}
    \label{fig:model}
\end{figure} 




\newpage
\tableofcontents
\newpage

\section{Introduction}
\begin{figure}[!t]
    \centering
    \includegraphics[width=\linewidth]{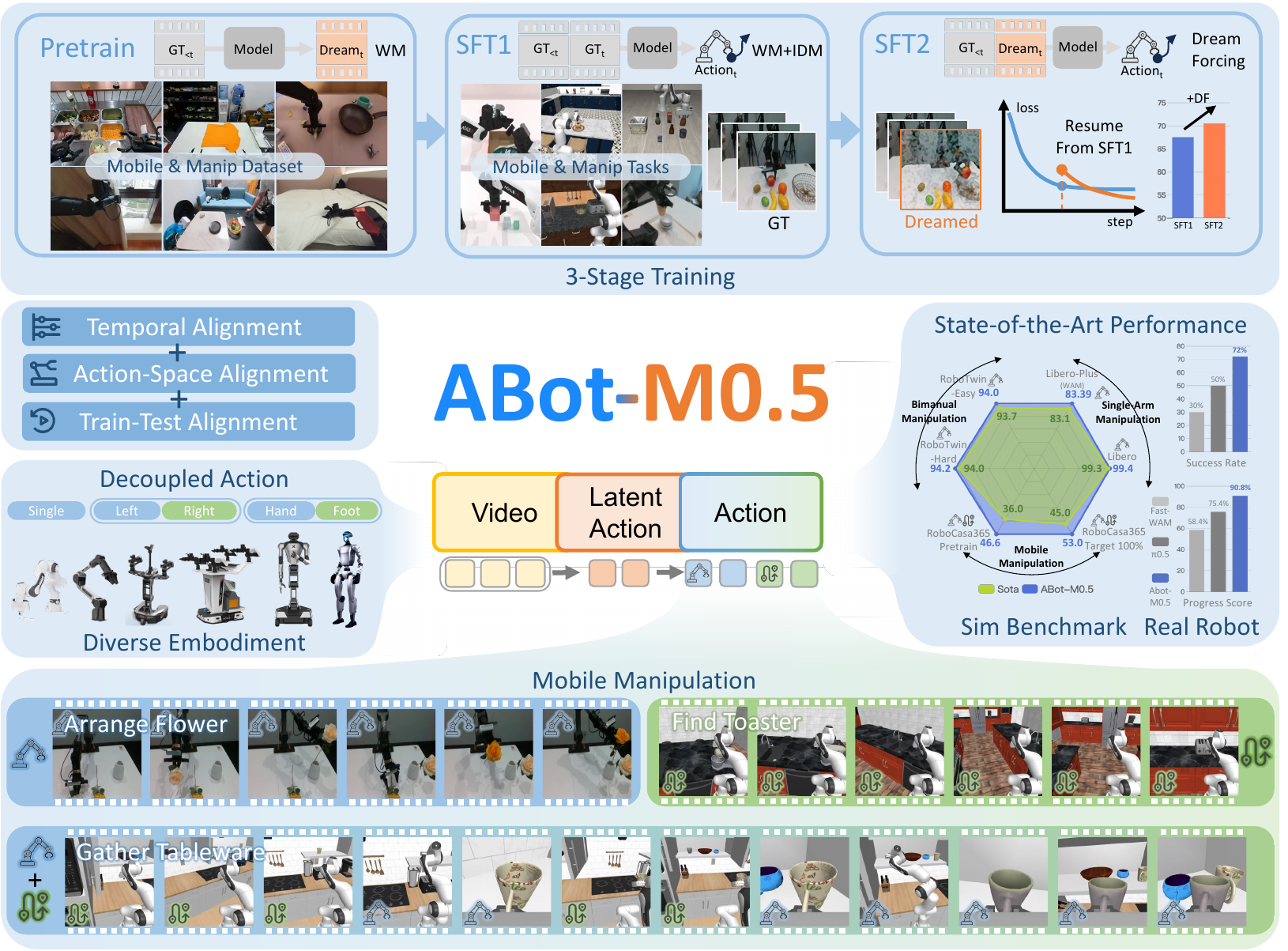}

    \caption{\textbf{Overview of ABot-M0.5.} ABot-M0.5 is a granularity-aligned, action-disentangled, and train-test-consistent world-action model for mobile manipulation. 
  \textbf{Top:} The 3-stage training pipeline (Pretrain, SFT1, SFT2 ) progressively improves the model. The final Dream-Forcing stage (SFT2) uses model-predicted videos to train inverse dynamics, significantly boosting robustness. 
  \textbf{Left \& Center:} To solve structural mismatches, we design a Video $\rightarrow$ Latent Action $\rightarrow$ Action pipeline for temporal alignment, and use a dual-level architecture to decouple different action spaces (e.g., base movement and arm manipulation), making the framework adaptable to diverse robot embodiments. 
  \textbf{Right:} ABot-M0.5 achieves state-of-the-art performance on both simulation benchmarks (radar chart) and real-world mobile manipulation tasks (bar charts). 
  \textbf{Bottom:} Real-world execution sequences (e.g., Arrange Flower, Find Toaster) show successful long-horizon mobile manipulation, where blue and green colors distinguish the navigation and manipulation phases.}
    \label{fig:teaser}
\end{figure}

Mobile manipulation is a defining capability for general-purpose robots: a capable embodied agent must not only manipulate objects, but also navigate through cluttered environments, maintain long-horizon task context, and execute precise interactions under changing viewpoints and scene dynamics \cite{szot2022habitat20traininghome,yenamandra2024homerobotopenvocabularymobilemanipulation,li2024behavior1khumancenteredembodiedai,huo2026abot}. Despite rapid recent progress in embodied AI, current embodied learning paradigms still fall short of this goal. Reactive Vision-Language-Action (VLA) policies lack explicit world modeling and long-horizon memory \cite{brohan2023rt2,kim2024openvla,black2026pi0visionlanguageactionflowmodel,intelligence2025pi05visionlanguageactionmodelopenworld,intelligence2026pi07steerablegeneralistrobotic}, while emerging world-model-based methods, though promising, are still not structured in a way that matches the demands of mobile manipulation \cite{ye2026worldactionmodelszeroshot,bi2025motusunifiedlatentaction,ha2018worldmodels,hafner2024dreamerv3}.

The central limitation is not merely insufficient model scale, but a structural mismatch between current world-action learning methods and the requirements of mobile manipulation. Effective mobile manipulation requires alignment at three levels. First, \textbf{temporal granularity} must be aligned, since coarse video prediction must ultimately support fine-grained, step-level control; otherwise, subtle but crucial dynamics such as contact onset, grasp closure, and alignment correction are easily blurred or lost \cite{hafner2024dreamerv3}. Second, \textbf{action space} must be aligned, since navigation and manipulation follow fundamentally different dynamics---for example, low-frequency global base motion versus high-frequency local arm control---yet are often optimized within a single entangled action space, leading to interference and suboptimal specialization \cite{brohan2023rt1,brohan2023rt2}. Third, \textbf{train-test consistency} must be aligned, since actions at deployment time are conditioned on the model's own predicted future observations rather than ground-truth futures, creating train-test mismatch and compounding errors over long autoregressive rollouts \cite{ye2026worldactionmodelszeroshot,li2026causalworldmodelingrobot,ha2018worldmodels,hafner2024dreamerv3}.

\ABotWAM{} is designed around this alignment principle. Rather than addressing these issues as isolated patches, it provides a unified framework for alignment-aware world-action learning in mobile manipulation. To align temporal granularity, \ABotWAM{} introduces frame-level latent actions between video latents and robot actions, forming a fine-grained intermediate action space that captures local visual state transitions and is less tied to any specific embodiment. This factorizes the direct video-to-action mapping into a video-to-latent-action-to-action pipeline, allowing the model to recover motion intent from coarse visual dynamics and translate it into executable low-level control. To align action structure, \ABotWAM{} adopts a dual-level Mixture-of-Transformers architecture that disentangles both modality-specific representations and heterogeneous action subspaces, enabling base motion and arm manipulation to be modeled separately within a unified system. To align inference conditions, \ABotWAM{} further employs a Dream Forcing training strategy that exposes inverse dynamics learning to self-dreamed videos, directly aligning the conditioning context of training with that of inference, thus improving robustness to prediction errors.

Extensive experiments on challenging mobile manipulation and manipulation benchmarks show that \ABotWAM{} achieves strong performance in both long-horizon and fine-grained manipulation tasks. These results indicate that progress in mobile manipulation depends not only on scaling model capacity or data volume, but also on aligning world modeling, action abstraction, and deployment-time behavior within a single coherent framework \cite{ma2026vla_survey}. More broadly, they point to a practical path for extending WAMs from stationary manipulation to mobile manipulation.

Our main contributions are as follows:
\begin{itemize}
    \item We identify three core structural bottlenecks that limit existing WAMs in mobile manipulation: temporal granularity mismatch between coarse video prediction and fine-grained control, action structure mismatch between heterogeneous mobility and manipulation behaviors, and context mismatch between training and autoregressive inference.
    \item We propose \ABotWAM{}, a new WAM architecture for mobile manipulation that addresses these bottlenecks through intermediate latent actions, a dual-level Mixture-of-Transformers design, and Dream Forcing, enabling fine-grained motion abstraction, structured action decoupling, and train-test consistent inverse dynamics learning.
    \item We demonstrate strong results on challenging mobile manipulation and manipulation benchmarks, with clear gains in long-horizon task success and fine-grained manipulation accuracy, and validate the contribution of each component through extensive ablations.
\end{itemize}

\section{Alignment-Aware World-Action Learning}
\label{sec:alignment_wam}

This section formalizes mobile manipulation as an alignment-aware world-action learning problem. Rather than treating mobile manipulation as a simple long-horizon extension of stationary manipulation, we view it as a setting in which world modeling, action modeling, and deployment-time rollout must be jointly aligned. This perspective provides a unified explanation for why current embodied learning methods struggle in mobile manipulation, and motivates the design of \ABotWAM{} in the next section.

\subsection{Problem Setting}
\label{subsec:problem_setting}

We consider language-conditioned mobile manipulation tasks in which a robot must execute long-horizon behaviors by jointly performing navigation and object interaction in visually complex environments 
\cite{szot2022habitat20traininghome,yenamandra2024homerobotopenvocabularymobilemanipulation,li2024behavior1khumancenteredembodiedai}. At each time step $t$, the agent receives a language instruction $l$, a multi-view visual observation $o_t$, and optionally a history of past observations and actions. The goal is to generate low-level executable actions $a_t$ that complete the task over an extended horizon while remaining consistent with future world evolution.

Formally, let $o_t = \{I_t^{(1)}, \dots, I_t^{(N_c)}\}$ denote the multi-view observation at time $t$, where $N_c$ is the number of cameras and $I_t^{(i)}$ is the image captured by the $i$-th camera. Given an observation history $o_{\le t}$, an action history $a_{<t}$, and a language instruction $l$, the policy aims to predict future behavior over a horizon $H$. In reactive policies, this is often formulated as a direct conditional mapping over a chunk size $H$~\cite{brohan2023rt1,kim2024openvla,black2026pi0visionlanguageactionflowmodel,Diffusionpolicy}:
\begin{equation}
a_{t:t+H-1} \sim \pi(\cdot \mid o_{\le t}, a_{<t}, l).
\end{equation}
Such a formulation is effective when the required action mainly depends on the current observation and the temporal horizon is short. However, it becomes increasingly brittle in mobile manipulation, where future decisions depend not only on the current state, but also on how the environment is expected to evolve under the robot's own future actions.

World Action Models (WAMs) address this limitation by jointly modeling future observations and future actions \cite{ye2026worldactionmodelszeroshot,bi2025motusunifiedlatentaction,yuan2026fastwamworldactionmodels,ye2026gigaworldpolicyefficientactioncenteredworldaction,li2026causalworldmodelingrobot}. Let $z_{t+1:t+H}$ denote the compressed video latent of the future observations over the time horizon $t+1:t+H$. Instead of directly predicting actions from the current observation, a WAM models a structured future trajectory:
\begin{equation}
(z_{t+1:t+H}, a_{t:t+H-1}) \sim p(\cdot \mid o_{\le t}, a_{<t}, l).
\end{equation}
This formulation introduces explicit future world modeling and provides a natural interface for long-horizon rollout, since future prediction and action prediction are embedded in the same autoregressive process.

However, directly applying existing WAM formulations to mobile manipulation remains insufficient. Mobile manipulation differs from stationary manipulation in at least three ways. First, future world evolution spans larger viewpoint changes and more diverse scene transitions due to robot movement. Second, the action space becomes heterogeneous, since both mobility and manipulation must be generated within a single policy. Third, rollout robustness becomes more important, because long-horizon mobile tasks amplify small prediction errors over time.

These differences suggest that mobile manipulation should not be viewed merely as a larger version of stationary manipulation. Instead, it should be treated as a world-action learning problem with stricter structural requirements on representation, control, and rollout.

To make this explicit, the core challenge in mobile manipulation lies in bridging two fundamentally different spaces: the coarse, long-term future video latents $z_{t+1:t+H}$ that capture global world evolution, and the fine-grained, heterogeneous executable robot actions $a_{t:t+H-1}$. Directly mapping the video latent to executable robot action is notoriously difficult due to the severe granularity and semantic gaps between them. 
For notational simplicity in the following text, we will abstract away the explicit horizon $H$ and denote $z_{t+1{:}t+H}$ and $a_{t{:}t+H-1}$ simply as $z_{t+1}$ and $a_t$, respectively.

Ideally, a successful mobile manipulation policy should learn a coherent hierarchical process: it must first anticipate how the visual world will evolve ($z_{t+1}$), then distill this macroscopic evolution into frame-level intermediate motion intents that capture local visual state transitions, and finally ground these intents into embodiment-specific low-level controls ($a_t$). Formally, this desired hierarchy can be conceptualized as:
\begin{equation}
\text{Video Latent } z_{t+1} \rightarrow \underbrace{\text{Frame-level Motion Intents}}_{\text{Bridging Space}} \rightarrow  \text{Robot Action } a_{t},
\end{equation}
where the intermediate bridging space serves as the crucial link connecting future world dynamics with fine-grained physical execution. However, how to effectively define, learn, and align this intermediate space remains an open question. In the next section, we will address this by introducing \textit{latent actions} to instantiate this bridging space, along with tailored architectures and training strategies to achieve full alignment.

\subsection{Core Bottlenecks in Mobile Manipulation}
\label{subsec:core_bottlenecks}

Under the formulation above, the key limitation of existing methods is not merely insufficient model scale, but a structural mismatch between how current WAMs are trained and the requirements of mobile manipulation \cite{ye2026worldactionmodelszeroshot,bi2025motusunifiedlatentaction,yuan2026fastwamworldactionmodels}. We identify three core bottlenecks.

\paragraph{Temporal Granularity Mismatch}
Existing WAMs typically model future observations in temporally compressed chunks. This design is computationally efficient and suitable for long-horizon video prediction, but it creates a mismatch between the temporal granularity of world modeling and that of control generation. In practice, future video latents may summarize multiple frames within a chunk, whereas robot actions must often be generated at every frame or control step.

This mismatch is especially problematic in mobile manipulation, where fine-grained interactions determine success. Behaviors such as grasp closure, contact onset, object release, fine alignment, and local collision avoidance often unfold over very short temporal windows. When world modeling is performed only at a coarse chunk level, these local transitions may be smoothed out or omitted, making it difficult for the policy to recover the precise motion intent required for execution.

\paragraph{Action Structure Mismatch}
Mobile manipulation introduces a heterogeneous action space that differs substantially from the action space of stationary manipulation. The robot must control both global mobility and local manipulation, and these two forms of behavior obey very different dynamics. Base movement tends to be low-frequency, smooth, and globally oriented. Arm manipulation, by contrast, is higher-frequency, local, and sensitive to contact-rich dynamics. Treating them as a single entangled action space forces the model to optimize over conflicting patterns within one shared representation.

This mismatch has two consequences. First, it increases optimization difficulty. Gradients from mobility-dominated trajectories and manipulation-dominated trajectories may interfere, preventing the model from specializing to either mode effectively. Second, it weakens compositionality. In many mobile manipulation tasks, base control and arm control must coordinate while remaining structurally distinct.

\paragraph{Rollout Condition Mismatch}
A third bottleneck arises from the discrepancy between how inverse dynamics is trained and how actions are actually generated at inference time. During training, inverse dynamics is usually conditioned on ground-truth future observations or their latent representations. During inference, however, such ground-truth futures are unavailable. The model must instead act based on its own predicted visual rollouts, which inevitably contain noise, uncertainty, and sometimes severe errors such as blurring, object drift, or hallucinated content.

This train-test mismatch creates a form of exposure bias in world-action learning, related to the distribution-shift studied in sequence prediction and imitation learning \cite{bengio2015scheduledsamplingsequenceprediction,ross2011reductionimitationlearningstructured}. The inverse dynamics model is optimized under ideal future conditions, but deployed under imperfect self-generated futures. In long-horizon mobile manipulation, the discrepancy compounds over time and can eventually derail execution.

\paragraph{Summary}
These three bottlenecks reveal a shared pattern: current methods are insufficiently aligned with the structure of mobile manipulation. Coarse visual prediction is misaligned with fine control, entangled action learning is misaligned with heterogeneous robot behavior, and ground-truth-conditioned training is misaligned with autoregressive deployment. \ABotWAM{} is designed around this observation. The next section presents the full model, including latent actions for temporal alignment, a dual-level Mixture-of-Transformers for structured action modeling, and Dream-Forcing for rollout-aligned inverse dynamics learning.

\section{The \ABotWAM{} Model}
\label{sec:model}

This section details the architecture of \ABotWAM{}. Guided by the alignment perspective introduced in ~\Cref{sec:alignment_wam}, our model directly addresses the three structural bottlenecks in mobile manipulation: temporal granularity mismatch, action structure mismatch, and train-test condition mismatch. At a high level, \ABotWAM{} factorizes world-action learning into a hierarchical cascade: it first predicts future visual dynamics, refines them into frame-level motion intents, and finally generates embodiment-specific executable actions.

\subsection{Overall Architecture and Notation}
\label{subsec:overall_architecture}

\ABotWAM{} is a video-action World Action Model built upon the Wan2.2 video diffusion backbone \cite{wan2025wanopenadvancedlargescale}. Given a language instruction $l$ and a sequence of multi-view observations, it jointly models future video latents, frame-level latent actions, and executable robot actions within a unified generative framework.

At the perception stage, a 3D VAE compresses continuous video observations $o_t = \{I_t^{(1)}, \dots, I_t^{(N_c)}\}$ into compact spatiotemporal video latents $z_t$, while a text encoder (e.g., UMT5) maps $l$ into conditional features. Crucially, we introduce a frame-level latent action $m_t$ to capture local visual state transitions, serving as a bridging representation between coarse video latents and fine-grained control. The generation process follows a structured cascade over clean (noise-free) variables:
\begin{equation}
z_{t+1} \rightarrow m_t \rightarrow a_t,
\end{equation}
where $z_{t+1}$, $m_t$, and $a_t$ denote the clean future video latent, clean latent action, and clean executable action, respectively. This factorization decomposes direct video-to-action prediction into three distinct stages: world modeling, motion abstraction, and control generation.

\begin{table}[!t]
\centering
\caption{Main notation used in ABot-M0.5.}
\label{tab:main_notation}
\small
\setlength{\tabcolsep}{4pt}
\renewcommand{\arraystretch}{1.12}
\begin{tabularx}{\linewidth}{lX lX}
\toprule
\textbf{Symbol} & \textbf{Meaning} & \textbf{Symbol} & \textbf{Meaning} \\
\midrule

$t$  & Time step
& $o_t$ & Raw multi-view observation  \\

$I_t$ & Raw frame 
& $z_{t+1}$ & Video latent \\

$m_t$ & Frame-level latent action
& $a_t$ & Executable robot action \\

$a_t^{\mathrm{move}}$ & Mobility action
& $a_t^{\mathrm{manip}}$ & Manipulation action \\

$H$ & Prediction horizon
& $N_c$ & Number of cameras \\

$X_t$ & Token 
& $\hat{z}_{t+1}$, $\hat{m}_t$,$\hat{a}_t$ & Dreamed latents \\

$l$ & Language instruction
& $\tilde{z}_{t+1}$, $\tilde{m}_t$, $\tilde{a}_t$ & Noisy latents \\

$\tau$ & Diffusion time step ($\tau \in [0,1]$)
& $p_z, p_m, p_a$ & Latent distributions \\

\bottomrule
\end{tabularx}
\end{table} 

To optimize this hierarchical cascade, we employ Conditional Flow Matching (CFM) as the unified generative objective across all stages. Taking the first stage (world modeling) as an example, given the ground-truth clean video latent $z_{t+1}$ and standard Gaussian noise $\epsilon \sim \mathcal{N}(0, I)$, we construct a conditional probability path at time step $\tau \sim \mathcal{U}(0,1)$. The video prediction objective is defined as:
\begin{equation}
\label{eq:cfm_loss_video}
\mathcal{L}_{\mathrm{z}} = \mathbb{E}_{z_{t+1}, \epsilon, \tau} \left[\left\| v_\theta^z\big(z_{t+1}^\tau; z_{<t+1}, m_{<t}, a_{<t}, \tau, l\big) - (z_{t+1} - \epsilon) \right\|_2^2 \right],
\end{equation}
where $z_{t+1}^\tau =\tau z_{t+1} +  (1-\tau)\epsilon$ is the interpolated state, and $v_\theta^z$ is the network regressing the target velocity field. Crucially, the conditioning context for $z_{t+1}$ strictly includes only historical states ($z_{\le t}, m_{<t}, a_{<t}$) and the language instruction $l$. The subsequent stages (predicting $m_t$ and $a_t$) follow analogous CFM objectives, but with progressively expanded receptive fields conditioned on the previously generated variables in the cascade (e.g., conditioning $m_t$ on the predicted $z_{t+1}$). This mathematically guarantees that the training-time information flow perfectly mirrors the autoregressive generation order used at inference.

Architecturally, the model processes three parallel token streams:
\begin{equation}
X_t = [X_{t+1}^z, X_t^m, X_t^a],
\end{equation}
where $X_{t+1}^z$, $X_t^m$, and $X_t^a$ correspond to video latent tokens, latent action tokens, and action tokens. To reflect the causal dependencies dictated by the above conditional objectives, we enforce an \textbf{asymmetric information flow}: video latent tokens ($X_{t+1}^z$) are masked from attending to latent action tokens ($X_t^m$), as future motions are inherently unknown during video prediction. Conversely, action tokens ($X_t^a$) explicitly attend to $X_t^m$, ensuring that final control is grounded in fine-grained motion intentions. 

To realize this alignment, the architecture integrates three key mechanisms, which are detailed in the subsequent subsections: 
(1) an \textbf{Intermediate Latent Action Modeling} module that bridges the temporal granularity gap; 
(2) a \textbf{Dual-level Mixture-of-Transformers (D-MoT)} that structurally decouples heterogeneous action subspaces (e.g., base mobility vs. arm manipulation) while maintaining modality-specific optimization; and 
(3) a \textbf{Dream Forcing Mechanism} that exposes the action stream to self-generated visual predictions, fundamentally resolving the train-test distribution shift.

\begin{figure}[t]
    \centering
    \includegraphics[width=\linewidth]{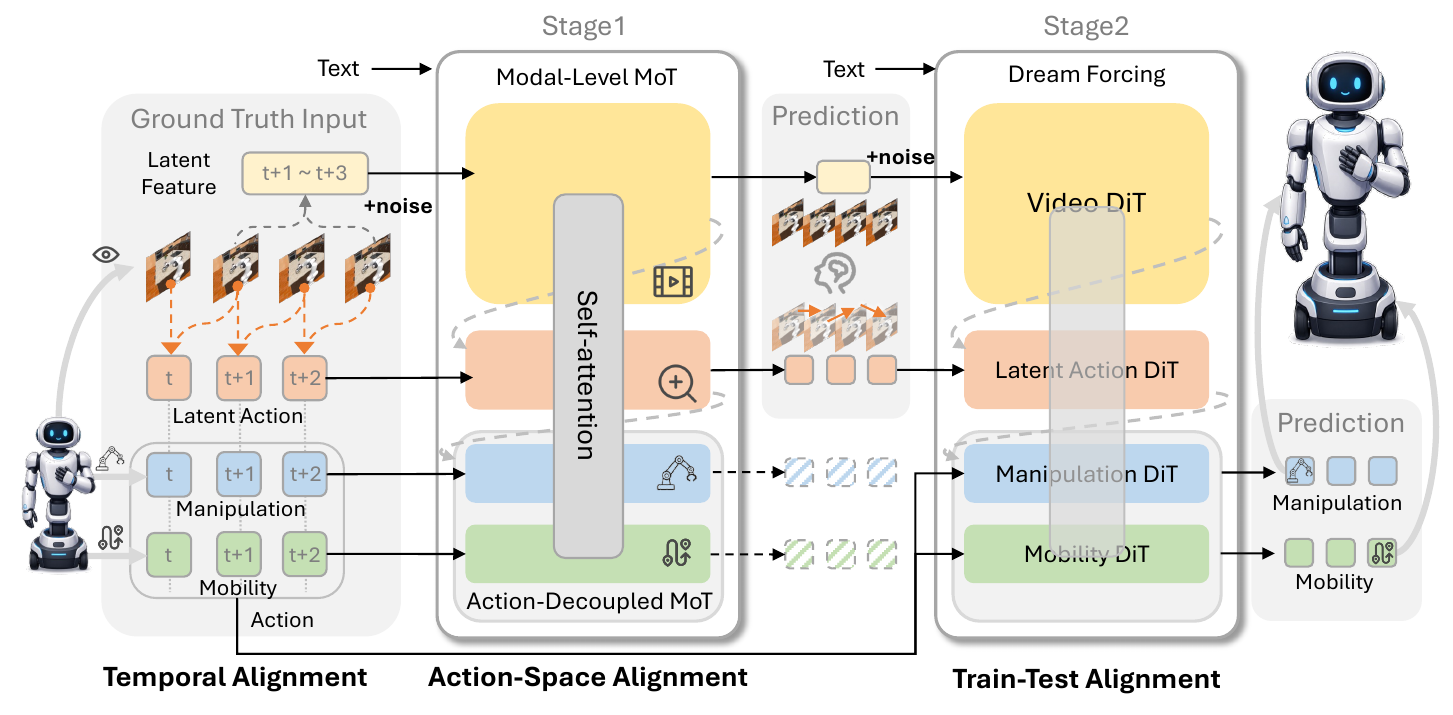}
    \caption{\textbf{Overall architecture of \ABotWAM{}.} The model jointly predicts future video latents, frame-level latent actions, and executable actions through a structured, asymmetric cascade design of a dual-level MoT. The Action-Decoupled MoT disentangled action into mobile and manipulation and predict together .}
    \label{fig:model_overview}
\end{figure}

\Cref{fig:model_overview} illustrates the overall architecture of \ABotWAM{} and \Cref{tab:main_notation} summarizes the core notation.
The remainder of this section elaborates on the design and implementation of these three core components.

\subsection{Intermediate Latent Action Modeling}
\label{subsec:latent_actions}





As established in ~\Cref{subsec:core_bottlenecks}, directly mapping coarse video latents to low-level actions creates severe temporal and structural mismatches. To resolve this, \ABotWAM{} introduces \textbf{frame-level latent actions} ($m_{0,t}$) as an intermediate, embodiment-agnostic representation. This design factorizes the generation process into a structured \textbf{three-stage cascade}:
\begin{equation}
\text{Context} \xrightarrow{\text{World Modeling}} z_{t+1} \xrightarrow{\text{Motion Abstraction}} m_{t} \xrightarrow{\text{Control Decoding}} a_{t}.
\end{equation}
Specifically, the model first predicts future video latents $z_{t+1}$ to capture macroscopic environmental evolution (Stage 1). It then refines these coarse dynamics into frame-level latent actions $m_{t}$ to represent fine-grained motion intents (Stage 2). Finally, it translates these intents into embodiment-specific executable actions $a_{t}$ (Stage 3). By explicitly modeling this hierarchy, the system decouples embodiment-agnostic physical priors from hardware-specific kinematics, enabling robust generalization across heterogeneous robot platforms.

\paragraph{Latent Action Extraction and Alignment}
A key advantage of latent actions is that they depend solely on visual state transitions, enabling extraction from large-scale, action-free video datasets without requiring robot kinematic labels \cite{ye2025latentactionpretrainingvideos,liang2025clamcontinuouslatentaction,tang2026alam}. Given consecutive frames $(I_t, I_{t+1})$, we utilize a frozen, pretrained latent action encoder $E_m$ to extract local motion representations:
\begin{equation}
m_{t} = E_m(I_t, I_{t+1}) \in \mathbb{R}^{d_m},
\end{equation}
where $d_m$ is the feature dimension. In multi-camera setups, we extract latent actions from each view and organize them into a unified spatiotemporal tensor. For a chunk with $H$ control steps and $N_c$ camera views, the aggregated latent action tensor is structured as $M=\{m_{t}^{view}\} \in \mathbb{R}^{H \times N_c  \times d_m}$.

Under this formulation, similar physical interactions (e.g., grasping an object) are mapped to proximate regions in the shared latent action space $\mathcal{M}$, regardless of the underlying robot morphology. This alignment is crucial for transferring physical priors across diverse embodiments.

\paragraph{Conditional Flow Matching for Latent Actions}
We formulate the generation of latent actions as a Conditional Flow Matching (CFM) problem, a simulation-free objective increasingly adopted in continuous-action robot policies \cite{lipman2023flowmatchinggenerativemodeling,black2026pi0visionlanguageactionflowmodel,pertsch2025fastefficientactiontokenization,intelligence2025pi05visionlanguageactionmodelopenworld}. 

Given the ground-truth clean latent action $m_t$ and standard Gaussian noise $\epsilon \sim \mathcal{N}(0, I)$, we construct a conditional probability path at time step $\tau \sim \mathcal{U}(0,1)$. The objective is defined as:
\begin{equation}
\label{eq:cfm_loss_la}
\mathcal{L}_{\mathrm{m}} = \mathbb{E}_{m_t, \epsilon, \tau} \left[\left\| v_\theta\big(m_t^\tau; z_{\le t+1}, m_{<t},a_{<t},\tau,\,l\big) - (m_t-\epsilon) \right\|_2^2 \right],
\end{equation}
where $m_t^\tau =\tau m_t +  (1-\tau)\epsilon$ is the interpolated state, and $v_\theta$ is the neural network regressing the target velocity field.

This loss supervises the model to synthesize high-fidelity latent actions through iterative denoising. By training exclusively on visual transitions conditioned on anticipated world dynamics, the model captures fine-grained, embodiment-agnostic motion intents that robustly guide the downstream control decoder.

\subsection{Dual-Level Mixture-of-Transformers}
\label{subsec:dual_level_mot}

While latent actions address the temporal granularity mismatch between world modeling and control, mobile manipulation also requires structured handling of heterogeneous action dynamics. To this end, \ABotWAM{} introduces a Dual-level Mixture-of-Transformers (D-MoT) architecture that disentangles heterogeneity at both the modality level and the action level.

\paragraph{Modality-Level Disentanglement}
The first level of D-MoT operates across the three parallel token streams: video latents ($X^z$), latent actions ($X^m$), and executable actions ($X^a$). Although these streams share the same Transformer trunk for cross-modal reasoning, they differ fundamentally in semantics and temporal roles. To prevent representational collapse, each modality is equipped with its own dedicated input projection, timestep embedding, and output head. This design ensures that the model maintains distinct representational spaces for world dynamics, motion intents, and hardware control, while still enabling flexible information exchange through shared self-attention layers.

\paragraph{Action-Level Disentanglement}
The second level of D-MoT operates strictly within the executable action stream ($X^a$). In mobile manipulation, the action vector $a_t$ inherently contains both mobility and manipulation dimensions, which exhibit distinct temporal frequencies and physical loss landscapes. Jointly predicting them with a single homogeneous head often leads to gradient interference, where the high-frequency manipulation signals dominate or destabilize the low-frequency mobility predictions.

\begin{figure}[t]
    \centering
    \includegraphics[width=0.5\linewidth]{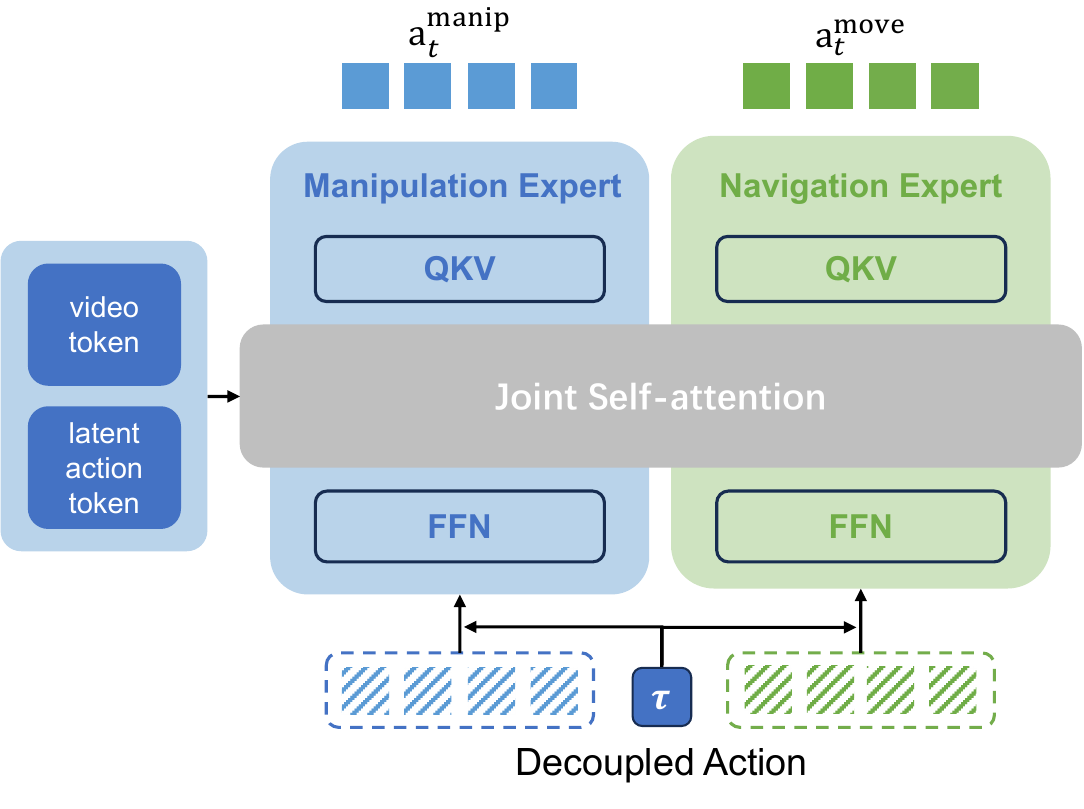}
    \caption{\textbf{Dual-level Mixture-of-Transformers.} The architecture disentangles modality-specific representations and heterogeneous action subspaces while preserving coordinated reasoning through shared attention. We omit the video and latent action expert for clarity.}
    \label{fig:dual_level_mot_model}
\end{figure}

To address this challenge, we explicitly decouple the action space $a_t$ into two distinct subspaces: manipulation $a_{t}^{manip}$ and mobility $a_{t}^{move}$. Specifically, as shown in~\Cref{fig:dual_level_mot_model}, we enforce a strict channel-to-subtower assignment, where each sub-tower is equipped with its own dedicated feed-forward network (FFN) and prediction head. This design enables each sub-tower to specialize in its dedicated action space, ensuring strictly decoupled learning dynamics between base mobility and arm manipulation.

\paragraph{Structured Joint Attention}
Despite the strict decoupling in the feed-forward layers, the model must still support coordinated reasoning between movement and manipulation (e.g., base repositioning directly affects grasp feasibility). To achieve this, D-MoT employs joint self-attention over the concatenated token streams at each layer. Under carefully designed causal and conditional masks, latent-action, mobility-action, and manipulation-action tokens participate in unified attention computation, while their subsequent FFN transformations remain branch-specific. This provides structured specialization without sacrificing cross-subspace coordination.

\paragraph{Subspace-Aware CFM Supervision}
We supervise the final action generation stage via conditional flow matching (CFM). To stabilize learning and prevent error accumulation across the cascade, we adopt a teacher-forced upstream conditioning protocol during training, where the action decoder receives ground-truth video latents $z_{\le k+1}$ and latent actions $m_{\le k}$. Furthermore, we make the noisy actions within the same chunk mutually visible to enable cross-subspace coordination. To maintain train-test consistency and improve inference efficiency, both subspaces share a single denoising timestep, aligning with the parallel joint denoising procedure at inference time and eliminating the need for separate noise schedules. 

Formally, at time $t$, the model predicts executable actions conditioned on the teacher-forced upstream representations and clean historical actions $a_{<t}=\{a_j^{\mathrm{move}}, a_j^{\mathrm{manip}}\}_{j<t}$. We sample a shared denoising timestep $\tau \in [0,1]$ for both action subspaces and independently draw Gaussian noises $\epsilon^{\mathrm{move}}, \epsilon^{\mathrm{manip}} \sim \mathcal{N}(0, I)$. The noisy mobility and manipulation actions are constructed as:
\begin{equation}
    a_t^{\mathrm{move},\tau} = \tau a_t^{\mathrm{move}} + (1-\tau)\epsilon^{\mathrm{move}}, \quad
    a_t^{\mathrm{manip},\tau} = \tau a_t^{\mathrm{manip}} + (1-\tau)\epsilon^{\mathrm{manip}}.
\end{equation}
The CFM objectives for the two branches are then defined as:
\begin{equation}
    \label{eq:cfm_loss_move}
    \mathcal{L}_{\mathrm{a}}^{\mathrm{move}} = 
    \mathbb{E}_{a_t^{\mathrm{move}}, \epsilon^{\mathrm{move}}, \tau}
    \left[
        \left\|
            v_{\theta}^{\mathrm{move}}
            \left(
                a_t^{\mathrm{move},\tau};
                z_{\le t+1}, m_{\le t}, a_{<t},
                a_t^{\mathrm{manip},\tau}, \tau, l
            \right)
            - \left(a_t^{\mathrm{move}} - \epsilon^{\mathrm{move}}\right)
        \right\|_2^2
    \right],
\end{equation}
\begin{equation}
    \label{eq:cfm_loss_manip}
    \mathcal{L}_{\mathrm{a}}^{\mathrm{manip}} = 
    \mathbb{E}_{a_t^{\mathrm{manip}}, \epsilon^{\mathrm{manip}}, \tau}
    \left[
        \left\|
            v_{\theta}^{\mathrm{manip}}
            \left(
                a_t^{\mathrm{manip},\tau};
                z_{\le t+1}, m_{\le t}, a_{<t},
                a_t^{\mathrm{move},\tau}, \tau, l
            \right)
            - \left(a_t^{\mathrm{manip}} - \epsilon^{\mathrm{manip}}\right)
        \right\|_2^2
    \right],
\end{equation}
where each branch takes the noisy action of the other branch as input, enabling cross-subspace coordination. 
The overall action objective is a weighted sum:
\begin{equation}
    \label{eq:loss_action}
    \mathcal{L}_{\mathrm{a}} = 
    \lambda_{\mathrm{move}} \mathcal{L}_{\mathrm{a}}^{\mathrm{move}}
    + \lambda_{\mathrm{manip}} \mathcal{L}_{\mathrm{a}}^{\mathrm{manip}},
\end{equation}
where $\lambda_{\mathrm{move}}$ and $\lambda_{\mathrm{manip}}$ balance the contributions of the two subspaces. Together, this subspace-aware CFM supervision closes the final stage of the $z_{t+1} \rightarrow m_t \rightarrow a_t$ cascade. By grounding action denoising in both anticipated world dynamics and fine-grained motion intents while maintaining cross-subspace information flow, the model is encouraged to produce temporally coherent and physically plausible action chunks.

\subsection{Dream Forcing for Train-Test Aligned Action Prediction}
\label{subsec:dream_forcing_model}


\begin{figure}[t]
    \centering
    \includegraphics[width=0.9\linewidth]{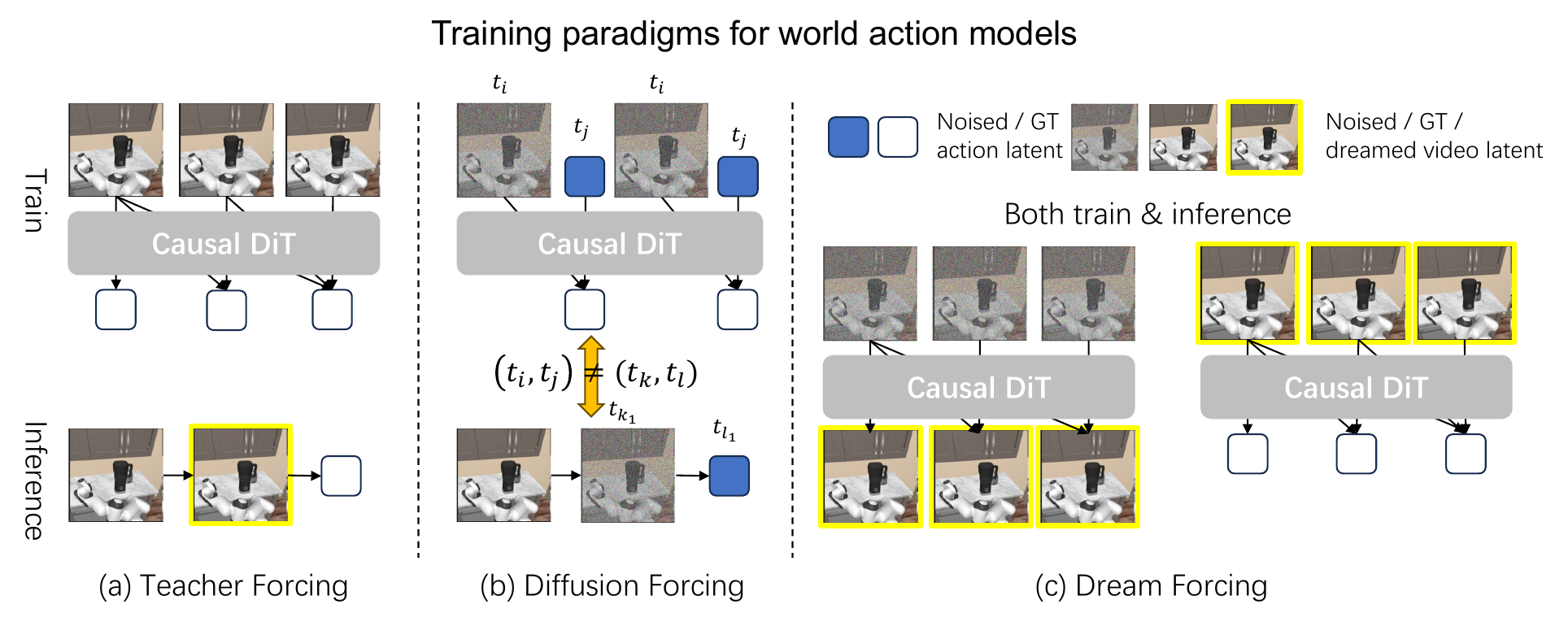}
    \caption{\textbf{Training paradigms for World Action Models.} Existing WAM training paradigms (a) and (b) suffer from train-test mismatch, as they condition on data distributions absent during inference. \textbf{(a)} In Teacher Forcing, the model is trained to denoise actions conditioned on clean, ground-truth future videos. \textbf{(b)} In Diffusion Forcing, the model jointly denoises future videos and actions with varying timesteps. Although the joint denoising scheme can be adapted for inference, it is difficult to mirror the exact timestep compositions used during training, thereby increasing the learning complexity and exacerbating the distribution gap. \textbf{(c)} Our Dream Forcing conditions action prediction on self-dreamed videos generated by the model itself. This paradigm closely mirrors the inference process, achieving faithful train-test alignment and bridging the distribution gap inherent in prior methods.
}
    \label{fig:dream_forcing}
\end{figure}




\paragraph{Train–Test Gap in Action Prediction}
Existing world action models (WAMs) mainly follow two training paradigms. The first is teacher-forcing paradigm~\cite{li2026causalworldmodelingrobot,pai2025mimic,feng2025vidar}, where action tokens are denoised conditioned on ground-truth video tokens. The second is diffusion forcing~\cite{ye2026worldactionmodelszeroshot, zhu2025unified,bi2025motusunifiedlatentaction}, where video and action tokens are jointly denoised within a unified diffusion process.

Despite their effectiveness, we identify a fundamental train-test gap in both paradigms, particularly for action prediction, as illustrated in \Cref{fig:dream_forcing}. In Teacher Forcing (\Cref{fig:dream_forcing}(a)), the action model is trained with access to clean GT video latents~\cite{li2026causalworldmodelingrobot, pai2025mimic, feng2025vidar}. However, such GT latents are unavailable at inference time; the model must instead condition on self-generated video latents that inevitably contain prediction errors and visual artifacts. This mismatch leads to severe exposure bias~\cite{ning2024elucidating, schmidt2019generalization, huang2026self}, where the action predictor has never been trained to interpret or compensate for its own dreamed visual states, leading to substantially degradation in action generation.

Diffusion forcing (\Cref{fig:dream_forcing}(b)) partially mitigates this issue by exposing action prediction to \textit{noisy} video latents during training~\cite{ye2026worldactionmodelszeroshot, zhu2025unified,bi2025motusunifiedlatentaction}. However, it introduces another form of discrepancy. During training, video and action tokens may appear under diverse and independently sampled noise timesteps, whereas inference follows a specific denoising trajectory whose exact timestep composition is difficult to reproduce. As a result, the model must learn action prediction under a broad set of artificial noise configurations, increasing optimization complexity and still leaving a mismatch between the training distribution and the actual inference-time conditioning context.

To address these limitations, we propose \textit{Dream Forcing}, a novel training paradigm that directly aligns the conditioning context of training with that of inference. Instead of conditioning action tokens on GT video latents or arbitrarily noised latents, Dream Forcing trains the action predictor on self-dreamed video latents produced by the model itself, as shown in \Cref{fig:dream_forcing}(c). This design exposes the action model to the same type of imperfect visual states it will encounter at inference time, enabling it to learn robust action generation under model-induced prediction errors and thereby substantially reducing the train-test gap.

\paragraph{Two-Phase Forwarding Strategy}
To implement the Dream Forcing training paradigm, we depart from the conventional approach of jointly optimizing multimodal tokens in a single forward pass. Instead, we introduce a two-phase forward pass strategy that decouples the generation of dreamed latents (Phase A) from the optimization of action prediction (Phase B).

The goal of \textbf{Phase A} is to synthesize the dreamed conditioning latents. Unlike autoregressive forcing methods in video generation~\cite{huang2026self, liu2025rolling, zhu2026causal, cui2025self} that must sequentially roll out multiple future chunks, our closed-loop robotic setting only requires dreaming the \textit{latest} future chunk. This is because historical chunks are continually grounded in real-world ground-truth (GT) observations during deployment~\cite{li2026causalworldmodelingrobot}. Consequently, rather than relying on a time-consuming sequential rollout, we employ a parallel generation strategy that produces all required dreamed latents for a batch in a single forward pass. Furthermore, since standard multi-step diffusion sampling remains computationally prohibitive for on-the-fly latent generation during training, we follow Self Forcing~\cite{huang2026self} and adopt a few-step denoising procedure to ensure training efficiency.

In \textbf{Phase B}, the model executes a second forward pass to predict actions conditioned on the dreamed latents synthesized in Phase A. Specifically, this shifts the action predictive distribution from the standard Teacher Forcing formulation:
\begin{equation}
a_{t} \sim p_a(\cdot \mid z_{\le t+1}, m_{\le t}, a_{<t}, l),
\end{equation}
to our Dream Forcing formulation:
\begin{equation}
a_{t} \sim p_a(\cdot \mid \hat{z}_{t+1}, z{\le t}, \hat{m}_t, m{<t}, a_{<t}, l),
\end{equation}
where only the future conditioning latents $z_{t+1}, m_{t}$ are replaced with their self-dreamed counterparts $\hat{z}_{t+1}, \hat{m}_{t}$.
By exposing the action prediction model to imperfect, dreamed visual contexts, Dream Forcing effectively eliminates the train-test distributional gap.




\begin{figure}[t]
    \centering
    \includegraphics[width=\linewidth]{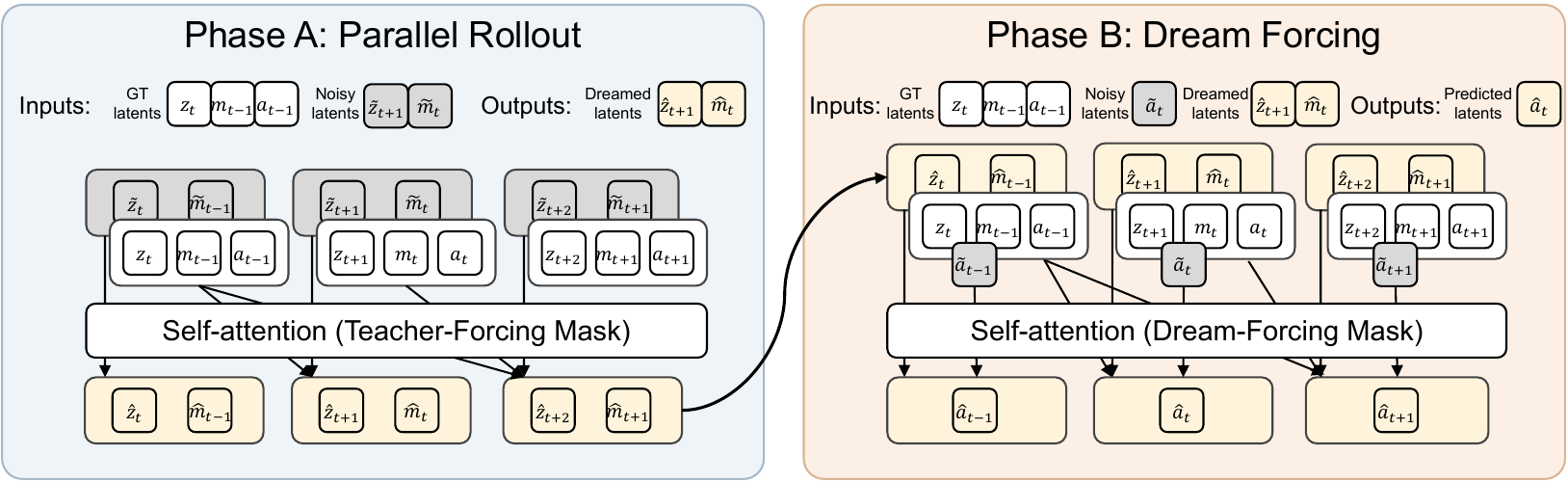}
    \caption{\textbf{Illustration of two-phase training strategy in Dream Forcing.} In \textbf{Phase A}, the model performs a standard forward pass to predict the velocity field, yielding clean dreamed latents $\hat{z}_{t+1}, \hat{m}_{t}$, which are sent to \textbf{Phase B} as conditions to conduct a second forward pass to predict the final action $\hat{a}_{t}$.}
    \label{fig:dream_forcing_mechanism}
\end{figure}

Taken together, these components form the core model design of \ABotWAM{}. Intermediate Latent Action Modeling address the temporal gap between world modeling and control, dual-level MoT addresses heterogeneity in modality and action structure, and Dream Forcing aligns inverse dynamics learning with autoregressive rollout. The next section describes how these components are trained progressively from large-scale pretraining to inference-aligned fine-tuning.

\section{Training Paradigm}
\label{sec:training}

To fully realize the representational and generative potential of \ABotWAM{}, we adopt a progressive training paradigm that moves from large-scale world modeling to rollout-consistent action learning. The overall strategy is motivated by the structure of the model itself. Since \ABotWAM{} jointly relies on future video prediction, frame-level latent action abstraction, and executable action generation, effective training must establish these capabilities in stages rather than optimizing them all from scratch under the most difficult rollout setting. We therefore organize training into three phases: large-scale pretraining for world modeling, self-supervised pretraining for latent action extraction, and progressive supervised fine-tuning for inverse dynamics learning. In addition, we introduce system-level optimizations to support efficient long-sequence video-action training.

\subsection{Pretraining Data}
\label{subsec:pretraining_data}

The pretraining corpus of \ABotWAM{} is constructed by combining large-scale public robot datasets with synthetic robotic data. The objective is to provide broad coverage over embodiments, environments, task structures, and manipulation dynamics, so that the model can acquire transferable priors before downstream fine-tuning.

All datasets are standardized into a unified data format for consistent processing, training, and evaluation. Following the data pipeline established in prior ABot work, we aggregate data from the following sources:

\begin{itemize}
    \item \textbf{OXE} \cite{embodimentcollaboration2025openxembodimentroboticlearning}: a large-scale multi-embodiment robotic dataset covering diverse scenes, tasks, and platforms. It provides broad visual and behavioral diversity and serves as a foundational source of embodied experience.
    \item \textbf{OXE-AugE} \cite{ji2025oxeauge}: an augmented extension of OXE designed to improve embodiment diversity, especially for single-arm morphologies.
    \item \textbf{Agibot-Beta} \cite{agibotworldcontributors2025agibotworldcolosseolargescale}: a high-quality dataset with structured task design, coherent action sequences, and long-horizon manipulation trajectories.
    \item \textbf{RoboCOIN} \cite{wu2026robocoinopensourcedbimanualrobotic}: a cross-embodiment dataset emphasizing dual-arm manipulation and hierarchical task structure.
    \item \textbf{RoboMind} \cite{wu2025robomindbenchmarkmultiembodimentintelligence}: a long-horizon manipulation dataset spanning both single-arm and dual-arm platforms, with strong cross-platform diversity.
    \item \textbf{Galaxea} \cite{jiang2025galaxeaopenworlddatasetg0}: a dataset with rich sensor signals and fine-grained sub-task annotations for complex, long-horizon manipulation. 
    \item \textbf{InternData-A1} \cite{tian2025interndataa1pioneeringhighfidelitysynthetic}: a large-scale synthetic robotic dataset constructed in simulation, covering diverse embodiments, manipulation skills, and scene configurations.
\end{itemize}

These datasets complement one another along several axes. OXE and OXE-AugE provide scale and embodiment diversity, Agibot-Beta and Galaxea provide higher-quality long-horizon task structure, RoboCOIN and RoboMind enrich cross-embodiment and dual-arm coverage, and InternData-A1 introduces synthetic scale and broader scene variation.
Galaxea also goes beyond arm-only operations by incorporating diverse base mobility tasks, where both are coordinated to complete long-horizon \textbf{mobile manipulation}.
Taken together, they enable \ABotWAM{} to learn robust visual dynamics and transferable motion priors before exposure to task-specific action supervision. We additionally include large-scale public robot corpora for diversity and robustness, including RoboNet \cite{dasari2020robonetlargescalemultirobotlearning}, BridgeData V2 \cite{walke2024bridgedatav2datasetrobot}, and DROID \cite{khazatsky2025droidlargescaleinthewildrobot}.

The same data infrastructure also supports latent action pretraining. Because frame-level latent actions depend only on visual frame pairs rather than control labels, large amounts of unlabeled or weakly labeled video can still contribute to motion abstraction learning. This broadens the effective supervision available to the model beyond standard action-annotated robotic datasets.

\subsection{World Model Pretraining}
\label{subsec:wm_pretraining}

The first phase of training focuses on pretraining the visual world model component of our framework. Initialized from the pretrained Wan2.2 5B weights \cite{wan2025wanopenadvancedlargescale}, the model is trained as an action-unconditioned future video predictor in an autoregressive (AR) manner. We perform full-parameter fine-tuning to adapt the Internet-scale spatiotemporal priors to robotic environments. This stage serves several critical purposes: it equips the model with robust scene and object representations, improves the quality of future latent predictions (which later serve as conditioning signals for action generation), and decouples visual world modeling from inverse dynamics learning, thereby significantly reducing the burden on downstream fine-tuning.

A key challenge in training on heterogeneous embodiment data is the significant semantic gap introduced by diverse camera configurations across different robot platforms and datasets, making it difficult for the video generation model to learn a consistent spatial representation when camera semantics are entangled. To address this, we introduce a fixed semantic slot allocation strategy. Specifically, we define four canonical video slots with predefined semantic roles: the first two slots are reserved for third-person views that capture the global scene and robot body configuration, while the last two slots are reserved for wrist-mounted views that provide fine-grained hand-object interaction details. For datasets that contain more than four camera views, we randomly sample a subset to fill the available slots, thereby introducing view-level data augmentation and preventing the model from overfitting to any particular camera arrangement. For datasets with fewer views than available slots, we apply zero padding to the unused slots.

To prevent zero-padded views from interfering with computation, missing views are represented as all-zero latent tensors and masked out in the self-attention layers, making them invisible to all valid views. The training objective is computed exclusively over valid views. Specifically, we optimize the visual world model using the conditional flow matching loss in the latent space, formulated as:
\begin{equation}
\label{eq:cfm_loss_pretrain}
\mathcal{L}_{\mathrm{z}}^{\mathrm{pretrain}} = \mathbb{E}_{z_t, \epsilon, \tau} \left[\left\| v_\theta^z\big(z_t^\tau; z_{<t}, \tau, l\big) - (z_t - \epsilon) \right\|_2^2 \right],
\end{equation}
where $v_\theta^z$ denotes the parameterized velocity field, $z_t^\tau$ is the interpolated latent at noise level $\tau$, $z_{<t}$ represents the autoregressive historical context, and $l$ is the conditioning input. We mask out this loss on padded regions so that no gradient propagates from artificial padding. This ensures the optimization is driven entirely by real visual observations and allows the model to seamlessly ingest multi-view data from heterogeneous robot embodiments with varying camera setups.

The pretrained world model is especially important in mobile manipulation. Compared with stationary manipulation, mobile tasks exhibit larger visual changes due to base movement, scene relocation, and changing camera viewpoints. A weak world model would therefore provide unstable or low-quality future predictions, severely limiting the effectiveness of downstream action learning. By pretraining world dynamics at scale, \ABotWAM{} starts fine-tuning from a much stronger representation of embodied future evolution.

\begin{figure}[t]
    \centering
    \includegraphics[width=\linewidth]{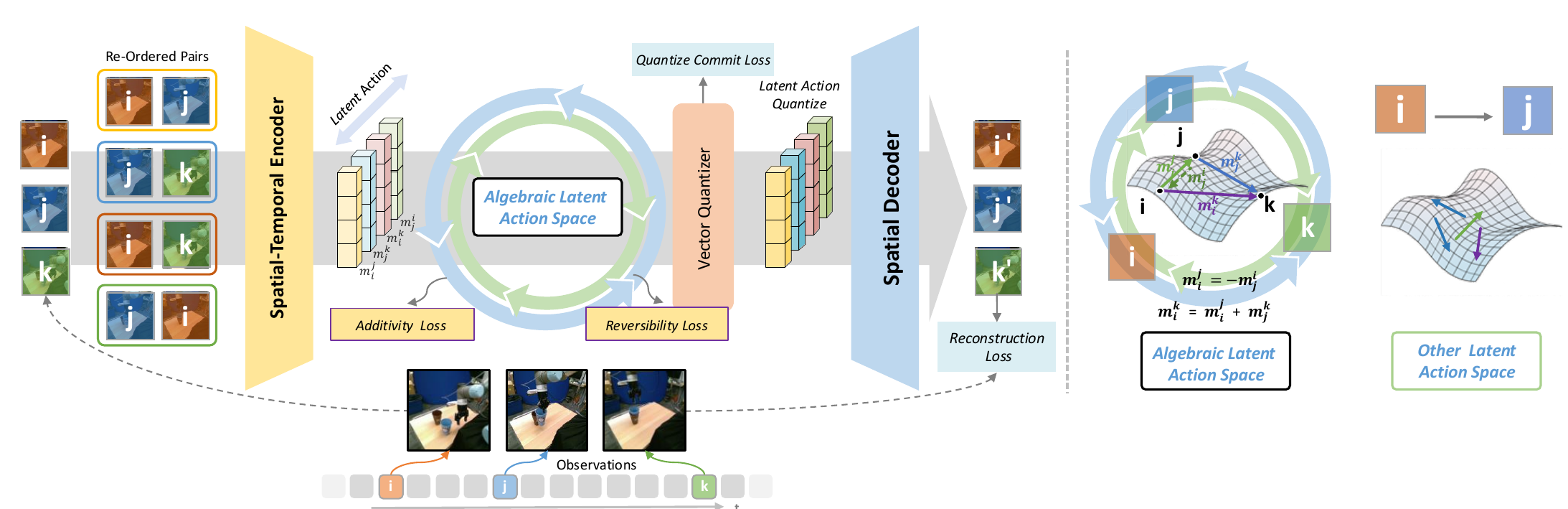}
    \caption{\textbf{Model structure of ALAM.} }
    \label{fig:alam}
\end{figure}

\subsection{Latent Action Model Pretraining}
\label{subsec:lam_pretraining}

The second phase pretrains the latent action encoder used to construct frame-level motion supervision. Unlike executable robot actions, latent actions are defined in terms of visual transitions between consecutive frames. This makes them suitable for self-supervised training on large-scale video data.

To obtain the latent action encoder $E_m$, we adopt the training framework proposed in ALAM \cite{tang2026alam}, shown in~\Cref{fig:alam}. Given a triplet of temporally ordered observations $(o_i, o_j, o_k)$ with $i < j < k$, the latent action model learns transition embeddings that capture how the observation changes over time. To encourage these embeddings to form a structured motion space, we impose algebraic consistency constraints over the learned transitions.

Specifically, let $m_i^j$ denote the latent action from $o_i$ to $o_j$. We enforce additive consistency:
\begin{equation}
\mathcal{L}_{\mathrm{add}} = \left\| m_i^k - (m_i^j + m_j^k) \right\|_2^2,
\end{equation}
which encourages longer temporal transitions to be approximately decomposable into shorter ones. We also enforce reversal consistency:
\begin{equation}
\mathcal{L}_{\mathrm{rev}} = \left\| m_i^j + m_j^i \right\|_2^2,
\end{equation}
which encourages the transition from $o_i$ to $o_j$ to be the inverse of the transition from $o_j$ to $o_i$.

In addition to these relational constraints, the model is trained with reconstruction and vector-quantization objectives to ensure that the latent code remains informative and compact. The full pretraining objective is motivated by structured latent-action learning and flow-based policy generation \cite{tang2026alam,lipman2023flowmatchinggenerativemodeling,tang2026one}:
\begin{equation}
\mathcal{L}_{\mathrm{LAM}} =
\lambda_{\mathrm{vq}} \mathcal{L}_{\mathrm{vq}} +
\lambda_{\mathrm{rec}} \mathcal{L}_{\mathrm{rec}} +
\lambda_{\mathrm{perc}} \mathcal{L}_{\mathrm{perc}} +
\lambda_{\mathrm{add}} \mathcal{L}_{\mathrm{add}} +
\lambda_{\mathrm{rev}} \mathcal{L}_{\mathrm{rev}}.
\end{equation}

After pretraining, we retain only the latent action encoder $E_m$ and discard the decoder and vector-quantization modules. The encoder is then frozen and used as an offline feature extractor to generate latent action labels for robot trajectories. In this way, the latent action supervision used by \ABotWAM{} is learned from large-scale visual motion rather than handcrafted from robot control signals.

This stage is crucial for two reasons. First, it provides a fine-grained intermediate representation that compensates for the temporal coarseness of future video latents. Second, it allows motion knowledge from unlabeled videos to be transferred into robotic action learning, thereby expanding the effective scope of pretraining beyond purely action-annotated robotic datasets.

\subsection{Progressive Supervised Fine-Tuning}
\label{subsec:progressive_sft}

After world model pretraining and latent action pretraining, \ABotWAM{} enters supervised fine-tuning on downstream robotic data. This stage introduces explicit action supervision and aligns the model with task-specific control distributions. Rather than directly training under the final rollout regime, we use a progressive strategy that first stabilizes world-action learning under clean future conditions and then gradually transitions to rollout-consistent conditioning.

\paragraph{Stage I: Joint World Model and Inverse Dynamics Fine-Tuning}
At the beginning of supervised fine-tuning, the pretrained world model still exhibits domain shift relative to the target datasets. Although it already encodes useful spatiotemporal priors, its future predictions are not yet sufficiently accurate to serve as reliable conditions for downstream action learning. If the inverse dynamics model were trained immediately on predicted futures, visual prediction errors would interfere with action learning before the model has adapted to the new data distribution.

We therefore begin with a stabilized fine-tuning stage in which latent action prediction and executable action prediction are conditioned on ground-truth future video latents. Under this setting, the model jointly predicts future video latents, latent actions, and executable actions:
\begin{align}
z_{t+1} &\sim p_z(\cdot \mid z_{\le t}, m_{<t}, a_{<t}, l), \\
m_{t} &\sim p_m(\cdot \mid z_{\le t+1}, m_{<t}, a_{<t}, l), \\
a_{t} &\sim p_a(\cdot \mid z_{\le t+1}, m_{\le t}, a_{<t}, l).
\end{align}
By aggregating~\Cref{eq:cfm_loss_la,eq:cfm_loss_video,eq:loss_action}, the total training loss is
\begin{equation}
\mathcal{L}_{\mathrm{SFT1}} =
\lambda_z \mathcal{L}_{\mathrm{z}} +
\lambda_m \mathcal{L}_{\mathrm{m}} +
\lambda_a \mathcal{L}_{\mathrm{a}}.
\end{equation}

This stage provides a stable environment for action learning. Since future visual conditioning is clean, the inverse dynamics model can focus on learning the mapping from future world evolution and latent motion intention to executable control, rather than compensating for noisy or inaccurate predictions. In effect, Stage I initializes the interaction between the world model and the action model under controlled conditions.

\paragraph{Stage II: Dream Forcing for Inference-Aligned Fine-Tuning}
Once the model reaches an initially converged regime, we transition to inference-aligned fine-tuning. At inference time, the model does not have access to ground-truth future video latents; instead, it must rely entirely on its own predicted future latents. To reduce this train-test discrepancy, we replace ground-truth future conditioning $z_{t+1}$ and $m_t$ with model-predicted future conditioning $\hat{z}_{t+1}$ and $\hat{m}_t$ when predicting action:
\begin{align}
a_{t} &\sim p_a(\cdot \mid \hat{z}_{t+1}, z_{\le t}, \hat{m}_{t},m_{<t}, a_{<t}, l).
\end{align}
And the ~\Cref{eq:cfm_loss_move,eq:cfm_loss_manip} are modified as:

\begin{equation}
\label{eq:cfm_loss_move_df}
\tilde{\mathcal{L}}_{\mathrm{a}}^{\mathrm{move}} = \mathbb{E}_{a_t^{\mathrm{move}}, \epsilon, \tau} \left[\left\| v_\theta^{\mathrm{move}}\big(a_t^{\mathrm{move},\tau}; \hat{z}_{\le t+1}, \hat{m}_{\le t}, a_{<t},a_t^{\mathrm{manip},\tau}, \tau, l\big) - (a_t^{\mathrm{move}} - \epsilon) \right\|_2^2 \right],
\end{equation}
and
\begin{equation}
\label{eq:cfm_loss_manip_df}
\tilde{\mathcal{L}}_{\mathrm{a}}^{\mathrm{manip}} = \mathbb{E}_{a_t^{\mathrm{manip}}, \epsilon, \tau} \left[\left\| v_\theta^{\mathrm{manip}}\big(a_t^{\mathrm{manip},\tau}; \hat{z}_{\le t+1}, \hat{m}_{\le t}, a_{<t},a_t^{\mathrm{move},\tau}, \tau, l\big) - (a_t^{\mathrm{manip}} - \epsilon) \right\|_2^2 \right].
\end{equation}

Then the overall loss becomes:
\begin{align}
\mathcal{L}_{\mathrm{SFT2}} &=
\lambda_z \mathcal{L}_{\mathrm{z}} +
\lambda_m \mathcal{L}_{\mathrm{m}} +
\lambda_a \tilde{\mathcal{L}}_{\mathrm{a}} \\ 
&= \lambda_z \mathcal{L}_{\mathrm{z}} +
\lambda_m \mathcal{L}_{\mathrm{m}} +
\lambda_a(\lambda_a^{\mathrm{move}}\tilde{\mathcal{L}}_{\mathrm{a}}^{\mathrm{move}} +
\lambda_a^{\mathrm{manip}} \tilde{\mathcal{L}}_{\mathrm{a}}^{\mathrm{manip}} )
\end{align}

This stage implements the dream-forcing mechanism introduced in Section~\ref{subsec:dream_forcing_model}. Future latent actions and executable actions are now predicted under model-generated futures rather than perfect future observations. As a result, the inverse dynamics model learns to tolerate video prediction noise, slight object drift, and imperfect future rollout conditions.

\begin{figure}[t]
    \centering
    \includegraphics[width=\linewidth]{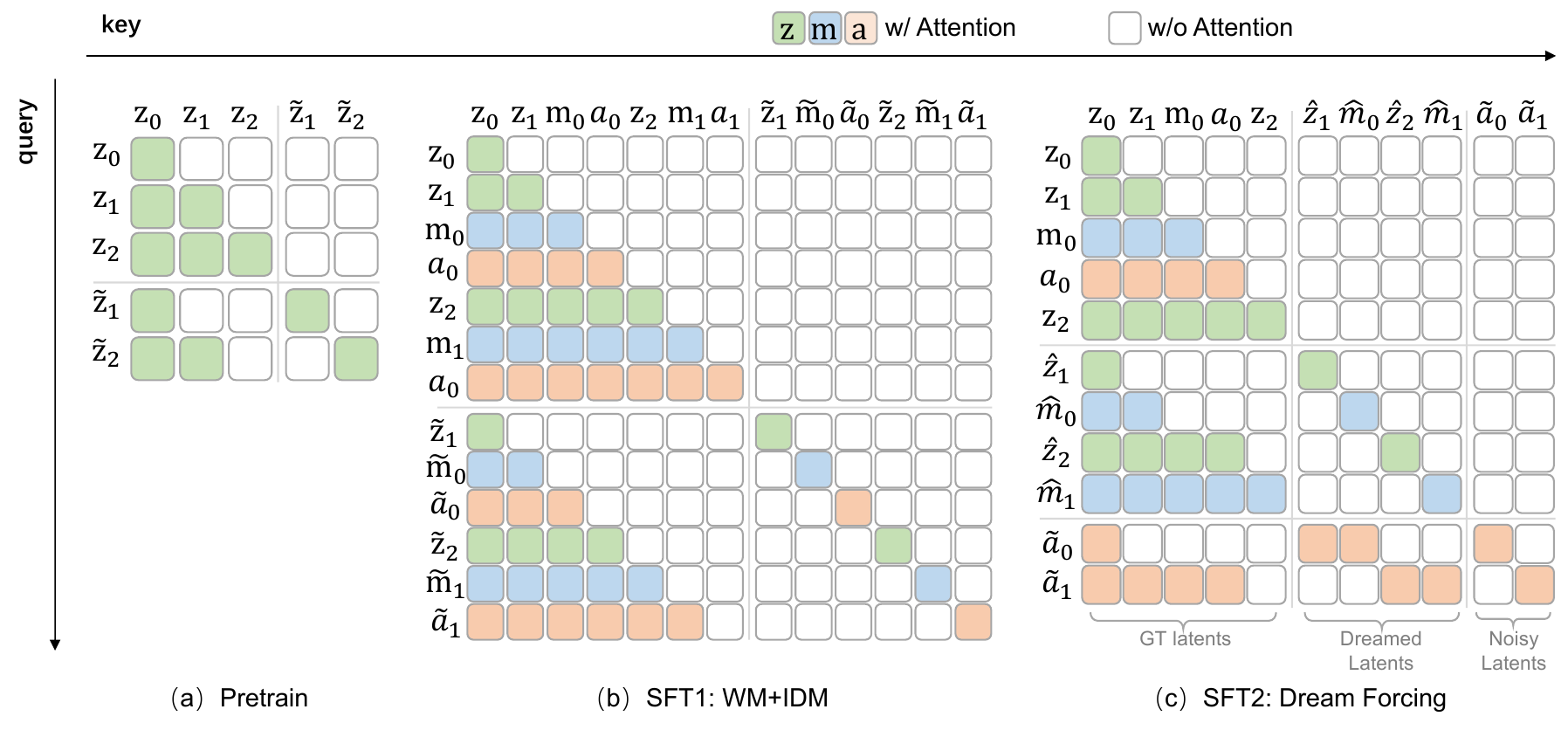}
    \caption{\textbf{Attention masks of different training stages.}}
    \label{fig:training_stages}
\end{figure}

\paragraph{Training Rationale}
The full fine-tuning strategy can be understood as a progressive alignment process. Stage I aligns the world model and action model on the downstream domain while shielding the inverse dynamics model from early prediction noise. Stage II then aligns the conditioning regime of action learning with the deployment regime of autoregressive rollout. In this way, the model is not merely optimized for one-step prediction accuracy, but for stable long-horizon performance under its own generated futures.

\subsection{Efficient Structured Attention and Latent Augmentation}
\label{subsec:efficient_attention}

The training pipeline described above requires long-sequence joint modeling over multiple token streams and multiple modalities. Without additional optimization, such training would be prohibitively expensive. We therefore introduce two practical techniques that improve efficiency and data utilization while preserving the intended semantics of the model.

\paragraph{Efficient Structured Attention}
\ABotWAM{} uses a structured sparse self-attention pattern to encode causal order, modality separation, and conditional visibility between token streams. A naive implementation with explicit block masks over the full attention matrix incurs substantial computational waste, especially for long sequences.

To address this, we reformulate the structured attention pattern as a set of dense sub-problems and implement them with variable-length FlashAttention \cite{dao2022flashattentionfastmemoryefficientexact}. Concretely, for each sample, frame, and token category, we precompute the valid ranges of query and key indices implied by the structured attention mask. These valid ranges are then packed into contiguous query-key-value segments, and multiple attention sub-problems are executed within a single variable-length FlashAttention kernel. This yields an implementation that is mathematically equivalent to the original structured attention pattern but much more GPU-efficient.

Compared with a FlexAttention-style baseline, this design substantially reduces kernel overhead, avoids unnecessary block padding, and lowers memory consumption. In practice, it provides approximately a $5\times$ speedup in the combined forward-backward pass for long-sequence video-action modeling.

\paragraph{Offset-Based Latent Augmentation}
To reduce the computational cost of repeatedly encoding raw video during training, we precompute video latent features at a fixed temporal stride $H$. A conventional implementation partitions a video starting from the first frame and always maps frames $[tH, (t+1)H]$ to the $t$-th latent feature. While simple, this rigid alignment effectively underutilizes the raw video and limits the diversity of latent segmentations seen during training.

We therefore introduce an offset-based indexing strategy. Instead of always starting from the first frame, we allow the starting offset $s$ to vary within
\begin{equation}
s \in \{0, 1, \dots, H-1\}.
\end{equation}
Under a given offset $s$, frames $[s+tH, ..., s+(t+1)H]$ are mapped to the $t$-th latent feature. Since there are $H$ possible offsets, this increases the number of valid latent segmentations by a factor of $H$. The resulting augmentation improves temporal diversity and empirically enhances robustness to small timing variations.

\paragraph{Role in the Full Training Pipeline}
These optimizations are not isolated engineering details. Efficient structured attention makes large-scale joint video-action training feasible under the multi-stream architecture of \ABotWAM{}, while offset-based latent augmentation improves data efficiency and temporal robustness. Together, they enable the progressive training pipeline to scale to long-horizon mobile manipulation without compromising the structural design of the model.

Taken together, the training paradigm of \ABotWAM{} combines large-scale world modeling, self-supervised motion abstraction, progressive action alignment, and efficient system-level optimization. This combination is critical to translating the architectural advantages of the model into practical performance on mobile manipulation benchmarks and real-world deployment. The next section evaluates this training strategy and the resulting model across a broad range of simulated and real-world tasks.

\section{Experiments}
\label{sec:experiments}

\begin{figure}[!h]
    \centering
    \includegraphics[width=1.0\linewidth]{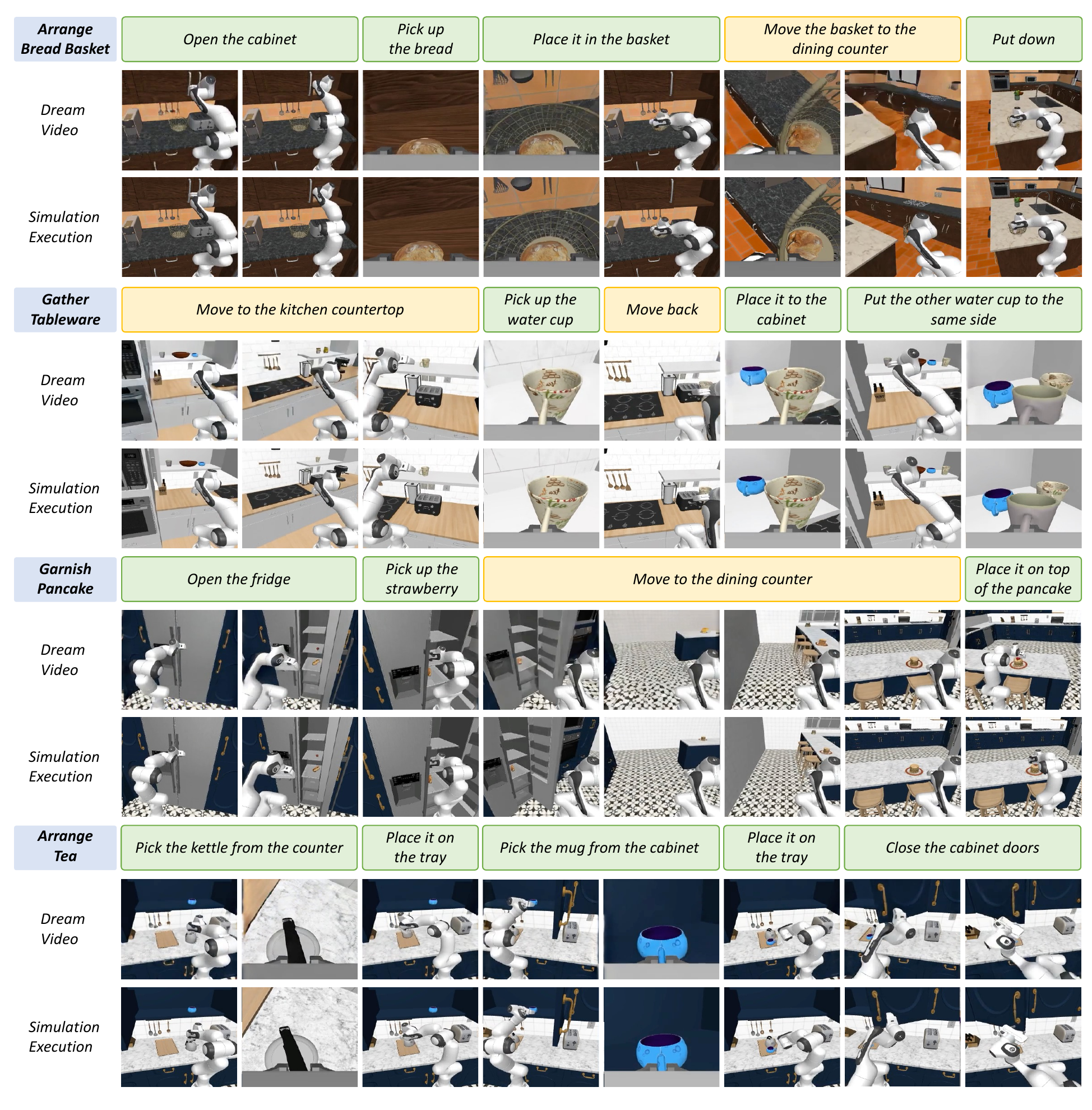}
    \caption{\textbf{Visualization of some results on RoboCasa365.} We show video frames of both real camera observations and model-dreamed scenarios for each sample. Here we decompose the task into several subtasks, using yellow and green to distinguish between mobility and manipulation, respectively.}
    \label{fig:subtask_robocasa}
\end{figure}

We evaluate \ABotWAM{} on a diverse set of mobile manipulation, manipulation, and real-world benchmarks to answer four questions. First, does the proposed framework improve long-horizon mobile manipulation performance over strong VLA and WAM baselines? Second, do the proposed architectural and training designs also benefit fine-grained manipulation beyond the mobile setting? Third, which components are responsible for the gains? Fourth, can the resulting model be deployed reliably in real-world scenarios? To answer these questions, we conduct experiments on RoboCasa365, RoboTwin 2.0, LIBERO / LIBERO-Plus, and real-world robotic tasks, together with detailed ablations and efficiency analysis.

\subsection{Experimental Setup}
\label{subsec:exp_setup}

\paragraph{Benchmarks}
We evaluate \ABotWAM{} on the following benchmarks.

\begin{itemize}
    \item \textbf{RoboCasa365} \cite{nasiriany2026robocasa365largescalesimulationframework}: a challenging mobile manipulation benchmark involving household tasks with both atomic and composite subtasks. It is the primary benchmark used to evaluate long-horizon mobile manipulation.
    \item \textbf{RoboTwin 2.0} \cite{chen2025robotwin20scalabledata}: a multi-task bimanual manipulation benchmark with both clean and randomized settings, used to evaluate generalization under scene variation.
    \item \textbf{LIBERO / LIBERO-Plus}~\cite{liu2023libero, fei25libero-plus}: compositional manipulation benchmarks that evaluate multi-task and long-horizon tabletop manipulation.
    \item \textbf{Real-World Tasks}: a set of real robotic tasks designed to test whether the learned world-action policy transfers beyond simulation.
\end{itemize}

Benchmark implementations follow the official setups of RoboCasa365, RoboTwin 2.0, and LIBERO/LIBERO-Plus.

\paragraph{Baselines}
We compare our method with prior Video-Language-Action (VLA) and World Action Models (WAMs). 
To ensure a comprehensive and objective evaluation, we carefully select baselines that encompass the most widely adopted, state-of-the-art, and most recent works, as well as those most closely related to our approach. Specifically, on the LIBERO-Plus benchmark, we extend our comparison beyond standard VLA models to include hybrid VLA+WM and pure WAM architectures, with a particular focus on highlighting the distinct advantages of our method over existing WAMs.

\paragraph{Metrics}
For each benchmark, we follow the official protocol and report benchmark-standard metrics. The primary evaluation metric is task success rate. For RoboCasa365, we additionally report performance across atomic seen, composite seen, and composite unseen task categories. For RoboTwin 2.0, we report average success rate under clean and randomized scenes across 50 tasks. For LIBERO and LIBERO-Plus, we report the standard benchmark success rate metric averaged across tasks. For real-world experiments, we evaluate on 5 tasks and report both success rate and process score. 

\paragraph{Implementation Details}
Unless otherwise specified, \ABotWAM{} is trained using the progressive pipeline described in Section~\ref{sec:training}. The video latent backbone, latent action encoder, and inverse dynamics components are jointly fine-tuned under the proposed training paradigm. For mobile manipulation, the action head models both mobility and manipulation controls. For non-mobile manipulation benchmarks, the same overall architecture is retained, with action routing specialized to the corresponding embodiment and control dimensions.

\subsection{Main Results on Mobile Manipulation}
\label{subsec:robocasa_results}

We first evaluate \ABotWAM{} on RoboCasa365, which serves as the primary benchmark for mobile manipulation. RoboCasa365 is particularly suitable for evaluating the proposed framework because many tasks require both long-horizon planning and precise manipulation, and because the benchmark contains diverse atomic and composite tasks under realistic household layouts.

\paragraph{Benchmark Protocol}
We follow the standard RoboCasa365 evaluation setup and report average performance together with category-specific results on atomic seen, composite seen, and composite unseen tasks. These categories capture increasingly challenging settings, from relatively localized interactions to longer task chains and unseen task compositions.

\paragraph{Main Comparison}
~\Cref{tab:robocasa_main} summarizes the comparison between \ABotWAM{} and representative baselines. \ABotWAM{} achieves strong overall performance, with particularly clear gains on long-horizon composite tasks. This is consistent with the design goals of the model: temporal alignment helps preserve fine-grained control-relevant dynamics, action decoupling improves structured coordination between mobility and manipulation, and Dream-Forcing improves robustness under long rollout.

\begin{table}[t]
\centering
\caption{\textbf{Evaluation results on RoboCasa365 Benchmark (pretraining).} \ABotWAM{} outperforms prior methods, achieving state-of-the-art performance. Furthermore, we introduce an enhanced \textit{condensed memory} mechanism, which yields further performance gains and sets a new record. A detailed discussion of this extension will be presented in our future work.}
\label{tab:robocasa_main}
\begin{tabular}{lcccc}
\toprule
\textbf{Method} & \textbf{Average} & \textbf{Atomic-Seen} & \textbf{Composite-Seen} & \textbf{Composite-Unseen} \\
\midrule
Diffusion Policy~\cite{Diffusionpolicy} & 6.1\% & 15.7\% & 0.2\% & 1.3\%  \\
$\pi_0$~\cite{black2026pi0visionlanguageactionflowmodel} & 14.8\% & 34.6\% & 6.1\% & 1.1\% \\
$\pi_{0.5}$~\cite{intelligence2025pi05visionlanguageactionmodelopenworld} & 16.9\% & 39.6\% & 7.1\% & 1.2\%  \\
GR00T-N1.5~\cite{nvidia2026gr00tn15} & 23.9\% & 50.7\% & 14.8\% & 2.7\% \\
GR00T-N1.6~\cite{nvidia2026gr00tn16} & 21.9\% & 51.1\% & 9.4\% & 1.7\% \\
GigaWorld-Policy 0.1~\cite{ye2026gigaworldpolicyefficientactioncenteredworldaction} & 20.7\% & 44.4\% & 11.8\% & 2.9\%  \\
RLDX-1~\cite{kim2026rldx} & 33.2\% & 63.0\% & 27.5\% & 5.4\%  \\
Qwen-RobotManip~\cite{yuan2026qwen-robotmanip} & 35.9\% & 68.6\% &  20.1\% & \textbf{14.9\%}  \\
Qwen-RobotManip-Context~\cite{yuan2026qwen-robotmanip} & 33.8\%  & 63.9\% & 22.6\% &11.2\%\\
\midrule
\rowcolor{lightorange}
\textbf{\ABotWAM{}~(Ours)} & 40.4\% &75.9\% & 38.3\% & 2.7\% \\
\rowcolor{lightorange}
\textbf{\ABotWAM{}}~(+Condensed Memory) & \textbf{46.6\%} & \textbf{79.4\%} & \textbf{48.3\%} & 7.9\% \\
\rowcolor{lightorange}

\bottomrule
\end{tabular}
\end{table}

\begin{table}[htbp]
  \centering
  \caption{\textbf{Evaluation results on RoboCasa365 benchmark (Target).} The suffixes -S and -U denote seen and unseen settings, respectively.}
  \label{tab:performance_comparison}
  \begin{tabular}{llcccc}
    \toprule
     \rowcolor{white}
     & \textbf{Method} & \textbf{Atomic-S} & \textbf{Composite-S} & \textbf{Composite-U} & \textbf{Average} \\
    \midrule
    \multirow{4}{*}{Target 100\%} & GR00T-N1.5~\cite{nvidia2026gr00tn15} & 60.6\% & 35.0\% & 33.3\% & 43.7\% \\
        & Fast-WAM~\cite{yuan2026fastwamworldactionmodels} & 59.1\% & 36.4\% &  33.2\% & 43.5\% \\
        & Lingbot-VA~\cite{li2026causalworldmodelingrobot}  & 63.5\% & 37.3\% & 32.1\% & 45.1\% \\
        \cmidrule{2-6}
        & \cellcolor{lightorange}\textbf{ABot-M0.5~(Ours)}  
        & \cellcolor{lightorange}\textbf{70.6\%} 
        & \cellcolor{lightorange}\textbf{44.3\%} 
        & \cellcolor{lightorange}\textbf{45.6\%} 
        & \cellcolor{lightorange}\textbf{54.2\%} \\
    \bottomrule
    \toprule
    \multirow{2}{*}{Target 10\%}  & GR00T-N1.5~\cite{nvidia2026gr00tn15} & 38.7\% & 11.0\% & 11.2\% & 21.0\% \\
        \cmidrule{2-6}
        & \cellcolor{lightorange}\textbf{ABot-M0.5~(Ours)}  & \cellcolor{lightorange}\textbf{49.0\%} & \cellcolor{lightorange}\textbf{23.4\%} & \cellcolor{lightorange}\textbf{15.4\%} & \cellcolor{lightorange}\textbf{30.1\%} \\
    \bottomrule
  \end{tabular}
\end{table}

\paragraph{Discussion}
The gains on composite tasks are especially important. Compared with stationary manipulation, composite mobile manipulation tasks expose the model to stronger viewpoint changes, longer action chains, and more severe error accumulation during rollout. The improvement of \ABotWAM{} in these settings suggests that the proposed framework is not merely improving low-level action prediction, but is better aligned with the structural requirements of long-horizon embodied control.

\paragraph{Qualitative Analysis}
\Cref{fig:subtask_robocasa} presents representative qualitative rollout examples on RoboCasa365. \ABotWAM{} is able to maintain coherent task progression across navigation and manipulation phases, while preserving fine-grained control during object interaction.

\subsection{Results on Manipulation Benchmarks}
\label{subsec:manip_benchmarks}

Although \ABotWAM{} is designed with mobile manipulation in mind, its core ideas are not limited to mobile settings. In particular, fine-grained latent action abstraction and rollout-aligned inverse dynamics learning should also benefit pure manipulation tasks. We therefore evaluate the model on RoboTwin 2.0 and LIBERO / LIBERO-Plus.

\paragraph{RoboTwin 2.0}
RoboTwin 2.0 evaluates multi-task bimanual manipulation under both clean and randomized scene settings. The clean setting tests execution under standard scene configurations, while the randomized setting evaluates robustness to background, layout, lighting, and tabletop variation. We adopt the standard training and evaluation protocol and report average task success rate.

\label{sec:exp_robotwin}

\begin{table}[htbp]
\centering
\caption{\textbf{Evaluation results on RoboTwin 2.0 benchmark.}}
\label{tab:model_comparison}
\begin{tabular}{lccc}
\toprule
\textbf{Model} & \textbf{Clean (Easy) } & \textbf{Randomized (Hard)} & \textbf{Average} \\ \midrule
X-VLA~\cite{xvla} & 72.80 & 72.84 & 72.82 \\
$\pi_{0.5}$~\cite{intelligence2025pi05visionlanguageactionmodelopenworld} & 82.70 & 76.80 & 79.75 \\
ABot-M0~\cite{yang2026abotm0vlafoundationmodel} & 86.06 & 85.08 & 85.57 \\
Qwen-VLA~\cite{wang2026qwenvlaunifyingvisionlanguageactionmodeling} & 86.10 & 87.20 & 86.65 \\
Fast-WAM~\cite{yuan2026fastwamworldactionmodels} & 91.90 & 91.80 & 91.85 \\
Lingbot-VA~\cite{li2026causalworldmodelingrobot} & 92.93 & 91.55 & 92.24 \\ 
HoloBrain-0~\cite{lin2026holobrain0technicalreport} & 91.90 & 92.30 & 92.10 \\
AttenA+~\cite{peng2026attenarectifyingactioninequality} & 93.10 & 91.90 & 92.50 \\
G0.5~\cite{galaxea2026g05} & 93.70 & 92.80 & 93.30 \\
Qwen-RobotManip~\cite{yuan2026qwen-robotmanip} & 93.70 & 94.00  & 93.85 \\
\midrule
\rowcolor{lightorange}
\textbf{ABot-M0.5 (Ours)} & \textbf{94.00} & \textbf{94.20}& \textbf{94.10} \\ \bottomrule
\end{tabular}
\end{table}

As shown in ~\Cref{tab:model_comparison}, \ABotWAM{} performs strongly under both clean and randomized settings. This indicates that the model does not rely solely on the mobile-specific structure of RoboCasa365, but also generalizes well to high-dimensional multi-task manipulation. In particular, the latent action abstraction appears to improve fine-grained control generation even when mobility is absent, while the rollout-aligned training contributes to robustness under visual variation.

\paragraph{LIBERO / LIBERO-Plus}
LIBERO and LIBERO-Plus focus on compositional tabletop manipulation and long-horizon multi-task generalization. These benchmarks test whether the model can preserve precise manipulation capabilities while maintaining broader compositional structure. 
\ABotWAM{} achieves leading or competitive performance on these benchmarks, indicating that the architecture does not sacrifice fine-grained manipulation performance when extended to mobile settings.
~\Cref{tab:libero_results} and ~\Cref{tab:libero-plus} report the main quantitative results.

\begin{table}[h]
\centering
\caption{\textbf{Evaluation results on LIBERO benchmark.}}
\label{tab:libero_results}
\small
\setlength{\tabcolsep}{8pt}

\begin{tabular}{lccccc}
\toprule
\textbf{Method} & \textbf{L-Spatial} & \textbf{L-Object} & \textbf{L-Goal} & \textbf{L-Long} & \textbf{Average} \\
\midrule
Diffusion Policy~\cite{Diffusionpolicy} & 78.5 & 87.5 & 73.5 & 64.8 & 76.1 \\
OpenVLA~\cite{kim2024openvla} & 84.7 & 88.4 & 79.2 & 53.7 & 76.5 \\
SpatialVLA~\cite{qu2025spatialvla}  & 88.2 & 89.9 & 78.6 & 55.5 & 78.1 \\
CoT-VLA~\cite{zhao2025cotvla}  & 87.5 & 91.6 & 87.6 & 69.0 & 83.9 \\
$\pi_0$-Fast~\cite{pertsch2025fastefficientactiontokenization}  & 96.4 & 96.8 & 88.6 & 60.2 & 85.5 \\
GR00T-N1~\cite{GR00T}  & 94.4 & 97.6 & 93.0 & 90.6 & 93.9 \\
$\pi_0$~\cite{black2026pi0visionlanguageactionflowmodel}  & 98.0 & 96.8 & 94.4 & 88.4 & 94.4 \\
F1~\cite{lv2025f1}  & 98.2 & 97.8 & 95.4 & 91.3 & 95.7 \\
InternVLA-M1~\cite{internvlam1}  & 98.0 & 99.0 & 93.8 & 92.6 & 95.9 \\
Discrete Diffusion VLA~\cite{liang2025discrete}  & 97.2 & 98.6 & 97.4 & 92.0 & 96.3 \\
$\pi_{0.5}$~\cite{intelligence2025pi05visionlanguageactionmodelopenworld} & 98.8 & 98.2 & 98.0 & 92.4 & 96.9 \\
GR00T-N1.6~\cite{nvidia2026gr00tn16} & 97.7 & 98.5 & 97.5 & 94.4 & 97.0\\
OpenVLA-OFT~\cite{OFT}  & 97.6 & 98.4 & 97.9 & 94.5 & 97.1 \\
Fast-WAM~\cite{yuan2026fastwamworldactionmodels} & 98.2 & \textbf{100.0} & 97.0 & 95.2 & 97.6 \\
Motus~\cite{bi2025motusunifiedlatentaction} & 96.8 & 99.8 & 96.6 & 97.6 & 97.7 \\
X-VLA~\cite{xvla} & 98.2 & 98.6 & 97.8 & 97.6 & 98.1 \\
ImageWAM~\cite{zhang2026imagewam} & 97.2 & 99.2 & 98.8 & 98.4 & 98.4 \\ 
Lingbot-VA~\cite{li2026causalworldmodelingrobot} & 98.5 & 99.6 & 97.2 & 98.5 & 98.5 \\ 
ABot-M0~\cite{yang2026abotm0vlafoundationmodel} & 98.8 & 99.8 & 99.0 & 96.6 & 98.6 \\
Being-H0.5~\cite{luo2026being05} & 99.2 & 99.6 & \textbf{99.4}& 97.4 & 98.9 \\
SaiVLA-0~\cite{shi2026saivla} & 99.8 &  \textbf{100.0} &	98.2 & 97.8 & 99.0 \\
PriorVLA~\cite{guo2026priorvla} & 99.4	& 99.8 &\textbf{99.4}& 97.6 & 99.1\\
Qwen-RobotManip~\cite{yuan2026qwen-robotmanip}  & - & - & - & - & 99.1 \\
Qwen-RobotManip-Context~\cite{yuan2026qwen-robotmanip}  & - & - & - & - & 99.2 \\
Being-H0.7~\cite{luo2026being07} & - & - & - & - & 99.2 \\
CORAL~\cite{luo2026coral} & 99.6 &	99.8 & 99.0	& \textbf{98.8} & 99.3 \\
\midrule
\rowcolor{lightorange}
\textbf{ABot-M0.5 (Ours)}  & \textbf{100.0} & 99.8 & \textbf{99.4} & 98.4 & \textbf{99.4} \\

\bottomrule
\end{tabular}

\label{tab:libero}
\end{table}

\begin{table}[h]\centering
\caption{\textbf{Zero-shot performance on LIBERO-Plus benchmark.} All methods are trained only on the standard LIBERO dataset without fine-tuning on the LIBERO-Plus dataset. Given the high sensitivity of WAMs to visual perturbations, we primarily compare against existing WAM models (best results in \textbf{bold}), where our method outperforms to achieve the overall SOTA. Previous VLA models are also included for reference, with their top results \underline{underlined}.}

\resizebox{\textwidth}{!}{
\begin{tabular}{lcccccccc}\toprule

\textbf{Method} & \textbf{Camera} & \textbf{Robot} & \textbf{Language} & \textbf{Light} & \textbf{Background} & \textbf{Noise} & \textbf{Layout} & \textbf{Total} \\
\midrule
\multicolumn{9}c{\textbf{VLAs}} \\
\midrule
OpenVLA~\cite{kim2024openvla} & 0.8 & 3.5 & 23.0 & 8.1 & 34.8 & 15.2 & 28.5 & 15.6 \\
OpenVLA-OFT~\cite{OFT} & 56.4 & 31.9 & 79.5 & 88.7 & 93.3 & 75.8 & 74.2 & 69.6 \\
OpenVLA-OFT\_w~\cite{OFT} & 10.4 & 38.7 & 70.5 & 76.8 & 93.6 & 49.9 & 69.9 & 55.8 \\
Openvla-OFT\_m~\cite{OFT} & 55.6 & 21.7 & 81.0 & 92.7 & 91.0 & 78.6 & 68.7 & 67.9 \\
NORA~\cite{hung2025nora} & 2.2 & 37.0 & 65.1 & 45.7 & 58.6 & 12.8 & 62.1 & 39.0 \\
WorldVLA~\cite{cen2025worldvla} & 0.1 & 27.9 & 41.6 & 43.7 & 17.1 & 10.9 & 38.0 & 25.0 \\
UniVLA~\cite{bu2025univla} & 1.8 & 46.2 & 69.6 & 69.0 & 81.0 & 21.2 & 31.9 & 42.9 \\
$\pi_0$~\cite{black2026pi0visionlanguageactionflowmodel} & 13.8 & 6.0 & 58.8 & 85.0 & 81.4 & 79.0 & 68.9 & 53.6 \\
$\pi_0$-Fast~\cite{pertsch2025fastefficientactiontokenization} & 65.1 & 21.6 & 61.0 & 73.2 & 73.2 & 74.4 & 68.8 & 61.6 \\
RIPT-VLA~\cite{RIPTVLA} & 55.2 & 31.2 & 77.6 & 88.4 & 91.6 & 73.5 & 74.2 & 68.4 \\
ABot-M0~\cite{yang2026abotm0vlafoundationmodel} & 60.4 & 67.9 & 86.4 & 96.2 & 91.6 & 86.4 & 82.6 & 80.5 \\
ACoT-VLA~\cite{zhong2026acot}  & 72.6 & 82.6 & 87.5 & 97.7 & 96.5 & 87.8 & \underline{88.1} & 86.6 \\
Qwen-RobotManip~\cite{yuan2026qwen-robotmanip} & 87.2 & 75.5 & 85.6 & 96.6 & 97.7 & 97.7 & 87.3 & 89.0 \\
Qwen-RobotManip-Context~\cite{yuan2026qwen-robotmanip} & \underline{89.9} & 83.9 & 86.5 & \underline{98.6} & \underline{99.9} & \underline{97.9} & 87.5 & \underline{91.4} \\
\midrule
\multicolumn{9}c{\textbf{VLA+WM}} \\
\midrule
VLA-JEPA~\cite{sun2026vla-jpea} & 63.3 & 67.1 & 85.4 & 95.6 & 93.6 & 66.3 & 85.1 & 79.5 \\
\midrule
\multicolumn{9}c{\textbf{WAMs}} \\
\midrule
Fast-WAM~\cite{yuan2026fastwamworldactionmodels} & 16.4 & 44.5 & 68.9 & 78.2 & 53.7 & 37.7 & 60.7 & 51.5 \\
Being-H0.7~\cite{luo2026being07} & - & - & - & - & - & - & - & 82.1 \\
Cosmos-Policy~\cite{kim2026cosmos} & 75.8 &63.3 & 81.7 & 96.5 & 88.9 & \textbf{92.7} & 82.2 & 82.2 \\
ImageWAM~\cite{zhang2026imagewam} & \textbf{80.8} & 50.3 & \underline{\textbf{91.4}} & \textbf{98.1} & 85.5 & 93.8 & 80.5 & 83.1 \\

\midrule
\rowcolor{lightorange}
\textbf{ABot-M0.5~(Ours)} & 70.5 &\underline{\textbf{87.4}} & 88.6 & 94.0 & \textbf{89.7} & 75.5 & \textbf{85.2} & \textbf{83.4} \\

\bottomrule
\end{tabular}
}

\label{tab:libero-plus}
\end{table}

\paragraph{Generalization beyond mobile manipulation}
These results are important for interpreting the proposed framework. They suggest that the benefits of \ABotWAM{} are not limited to a particular benchmark or to mobility-heavy settings. Instead, the gains appear to come from more general improvements in how future world evolution, local motion abstraction, and inverse dynamics are organized within the model.

\subsection{Ablation Studies}
\label{subsec:ablation}

We next perform ablation studies to understand the contribution of each major component. The ablations are designed to directly test the three alignment principles introduced earlier: temporal alignment, action structure alignment, and rollout alignment.

\begin{figure}[H]
    \centering
    \includegraphics[width=\linewidth]{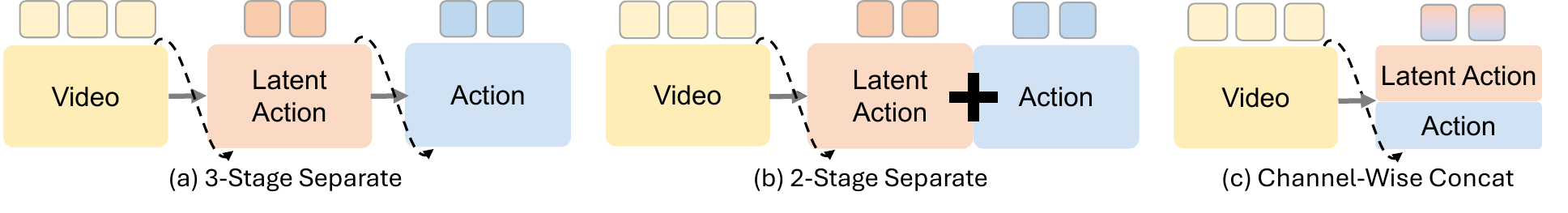}
    \caption{Different structures of latent action embeddings.}
    \label{fig:lam}
\end{figure}

\paragraph{Effect of Intermediate Latent Action}
To quantify the impact of Intermediate Latent Action Modeling, we compare our full framework against alternative configurations that either omit the latent action stream or generate actions directly from video latents. As shown in ~\Cref{fig:lam} and ~\Cref{tab:latent_action_ablations}, removing latent actions consistently degrades performance, particularly on tasks requiring precise manipulation and contact-rich interactions. This validates that coarse-grained video latents alone are insufficient for fine-grained control. When latent actions are integrated, the choice of training strategy plays a crucial role. The naive baseline, which directly maps video latents to actions yields a success rate of 87.60\%. Merging multi-view latent actions with actions along the channel dimension (Channel Concat) conflates distinct modality semantics under a shared denoising schedule, which limits the success rate to 91.06\%. Treating them as independent streams but sharing temporal indices without modal segregation (2-Stage Separate) allows unrestricted cross-attention, leading to representation leakage and a success rate of 90.86\%. In contrast, our 3-Stage Separate strategy assigns disjoint temporal and modal indicators to the video, latent action, and action streams, employing a structured attention mask to enforce strict causality. This formulation prevents video tokens from attending to latent action tokens, thereby eliminating leakage while allowing actions to condition on latent actions, thereby safeguarding modal boundaries and achieving a superior success rate of 94.0\%.

Under this 3-Stage Separate formulation, we further analyze the impact of conditioning dropout ($p_{\text{drop}}$) during training. While randomly masking latent actions is a standard practice to enable unconditional fallback, it introduces a train-test discrepancy under our cascade inference scheme where denoised latent actions are always available. Consequently, setting $p_{\text{drop}}=0.2$ degrades the success rate to 91.06\% due to an unstable conditioning signal during training. Conversely, setting $p_{\text{drop}}=0$ ensures perfect attention alignment between training and inference, allowing the action stream to reliably leverage the latent actions and boosting the success rate to 94.0\%. These results empirically validate that independent tokenization, hierarchical conditioning, and strict train-inference alignment are jointly essential for maximizing control precision.

\begin{table*}[htbp]
\centering
\begin{minipage}{0.48\textwidth}
\centering
\caption{\textbf{Ablation study of latent-action strategies} on RoboTwin 2.0 (Clean).}
\label{tab:latent_action_ablations}
\begin{tabular}{lcc}
\toprule
\rowcolor{gray!10} 
\textbf{Training Strategy} & \textbf{Drop} & \textbf{Success Rate} \\
\midrule
Baseline                   & -   & 87.60 \\
\midrule
2-Stage Separate           & 0   & 90.86 \\
2-Stage Channel Concat     & 0   & 91.06 \\
3-Stage Separate           & 0.2 & 91.06 \\
3-Stage Separate          & 0   & \textbf{94.00} \\
\bottomrule
\end{tabular}
\end{minipage}
\hfill
\begin{minipage}{0.48\textwidth}
\centering
\caption{\textbf{Ablation study on Dream Forcing (DF)} and different training stage of SFT on RoboCasa365 Target 100\%.}
\label{tab:ablation_df}
\begin{tabular}{lcc}
\toprule
\rowcolor{gray!10} Training Stage & Training Steps & Atomic-Seen \\
\midrule
SFT1 (Base) & 50k        & 67.55 \\
\midrule
SFT1            & +5k (55k)  & 66.78 \\
SFT1            & +10k (60k) & 68.90 \\
SFT2~(+DF)       & +5k (55k)  & \textbf{70.56} \\
\bottomrule
\end{tabular}
\end{minipage}
\end{table*}

\begin{figure}[H]
    \centering
    \includegraphics[width=\linewidth]{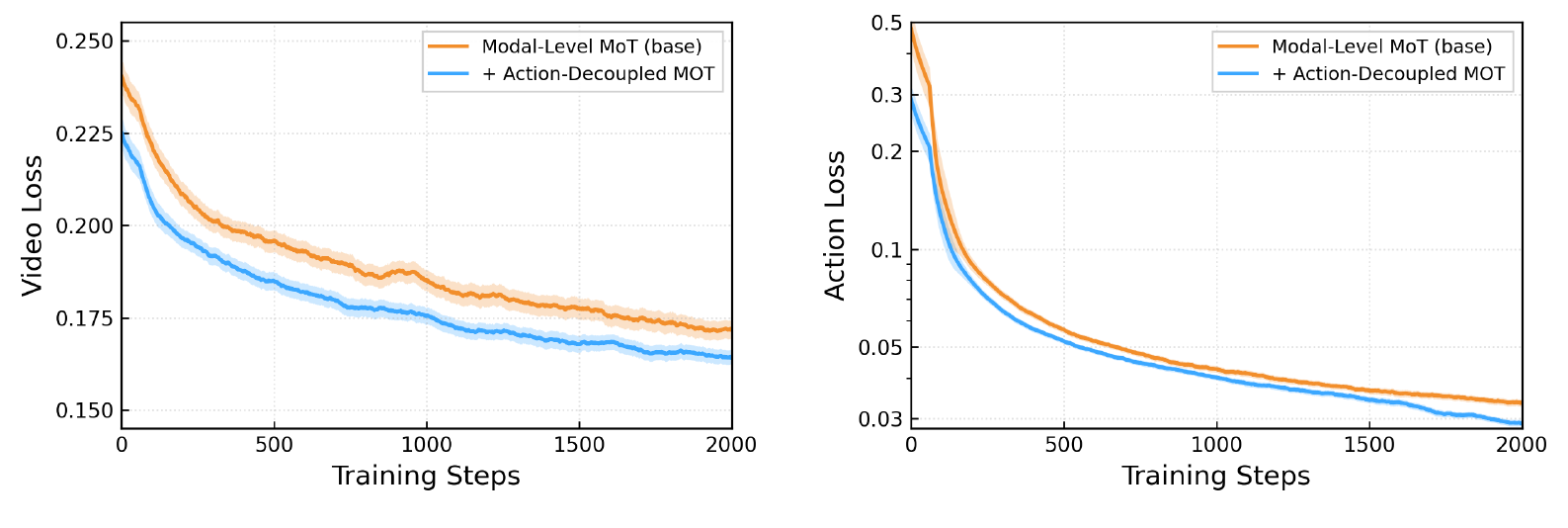}
    \caption{\textbf{Training dynamics of different MoT architecture designs.} Adding action-decoupled MoT demonstrates a faster convergence rate compared to the modal-level MoT.}
    \label{fig:mot_loss}
\end{figure}

\paragraph{Effect of Action-Decoupled MoT}
We evaluate the impact of the action-decoupled MoT on a selected subset of the RoboCasa365 Composite-Seen, which comprises long-horizon mobile manipulation tasks that demand frequent transitions between base navigation and arm manipulation. Specifically, we compare our action-decoupled MoT against the Modality-level MoT baseline that uses a single transformer for action modality prediction. Experimental results show that the action-decoupled MoT achieves $0.48$, outperforming the Modality-level MoT baseline which achieves $0.34$. Moreover, the training dynamics in~\Cref{fig:mot_loss} show that the action-decoupled MoT converges faster, demonstrating the effectiveness of action decoupling in reducing cross-action space gradient interference.

\paragraph{Effect of Dream Forcing}
To validate the efficacy of Dream Forcing, we conduct ablation studies on the RoboCasa365 target atomic seen subset, with quantitative results summarized in \Cref{tab:ablation_df}. We initialize our experiments from a shared warm-start checkpoint (optimized for 50k steps under the standard paradigm without Dream Forcing), which establishes a strong baseline with an average success rate of $67.55\%$.
Activating Dream Forcing and continuing training for mere 5k additional steps yields a substantial performance boost, elevating the success rate to $70.56\%$ (an absolute improvement of $3.01\%$ ). In contrast, continuing to optimize from the identical 50k-step checkpoint under the original teacher forcing manner leads to slight performance degradation, with the success rate dipping to 66.78 after 5k extra steps. Even after 10k additional steps, the success rate only recovers to $68.90\%$, still falling short of the $70.56\%$ achieved by Dream Forcing with only half the computational budget. These results demonstrate that Dream Forcing effectively bridges the train-test gap to boost the action prediction accuracy.

\begin{figure}[t]
    \centering
    \includegraphics[width=0.9\linewidth]{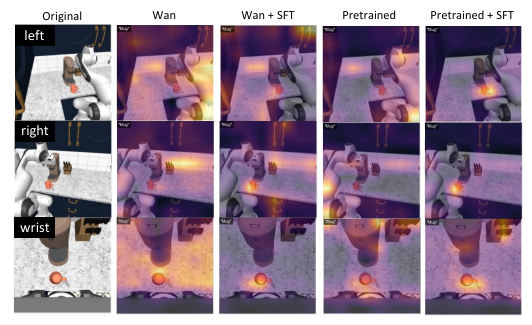}
    \caption{\textbf{Attention maps of different training strategies.} We calculate Text-video
    Text-conditioned attention maps under different training stages and model variants. Pretraining and downstream fine-tuning progressively improve the alignment between textual semantics and task-relevant visual regions.}
    \label{fig:Attentionmap}
\end{figure}

\paragraph{Effect of Pretraining and SFT}

We first quantitatively evaluate the impact of pretraining on the RoboCasa365 Atomic-Seen subset under the Target 10\% setting, which includes 16 training tasks with 50 trajectories randomly sampled per task. Using the same ABot-M0.5 architecture and SFT protocol, our pretrained model achieves a success rate of 49.0\% after fine-tuning. In contrast, a model initialized from Wan2.2 and directly fine-tuned on the same downstream data yields only 17.8\%, resulting in a large gap of 31.2\%. This gap indicates that SFT alone fails to learn enough visual dynamics and manipulation behaviors from limited demonstrations. Instead, our pretraining provides transferable visual and interaction priors, making downstream adaptation much more sample-efficient.

To investigate this improvement, we visualize text-conditioned attention maps for the word ``Mug'' across three camera views (see \Cref{fig:Attentionmap}). The original Wan model produces scattered attention and is often distracted by background clutter. After pretraining, this incorrect focus is reduced, and attention shifts toward the robotic arm and interaction regions, showing that the model has learned action-related visual priors. SFT further aligns these priors with task semantics. While Wan+SFT improves target activation, its attention remains unstable and is still distracted by other objects. In contrast, Pretrained+SFT produces the most focused and accurate attention, tightly concentrating on the target and its manipulation region across all views while ignoring irrelevant background objects.

Overall, pretraining and SFT work together to improve performance: pretraining establishes transferable visual and interaction priors, while SFT adapts them for downstream tasks. This combination is crucial in settings with limited data, where directly fine-tuning leads to poor task performance.

\paragraph{Summary}
Across all settings, each component plays a distinct and complementary role. Pretraining helps the network focus more on foreground interactive objects, while SFT enables it to locate target objects more accurately. The introduction of latent actions strengthens temporal alignment and the modality transition from video to actions, making the system more precise and stable in manipulation tasks. Meanwhile, the action-decoupled MoT separates different action spaces, which significantly improves performance in mobile manipulation tasks. Finally, the second-stage SFT, guided by Dream-Forcing, further boosts robustness, enabling the agent to perform more reliably and achieve better results during actual testing. Together, these components form a coherent design rather than a simple combination of independent tricks.

\begin{figure}[h]
    \centering
    \includegraphics[width=\linewidth]{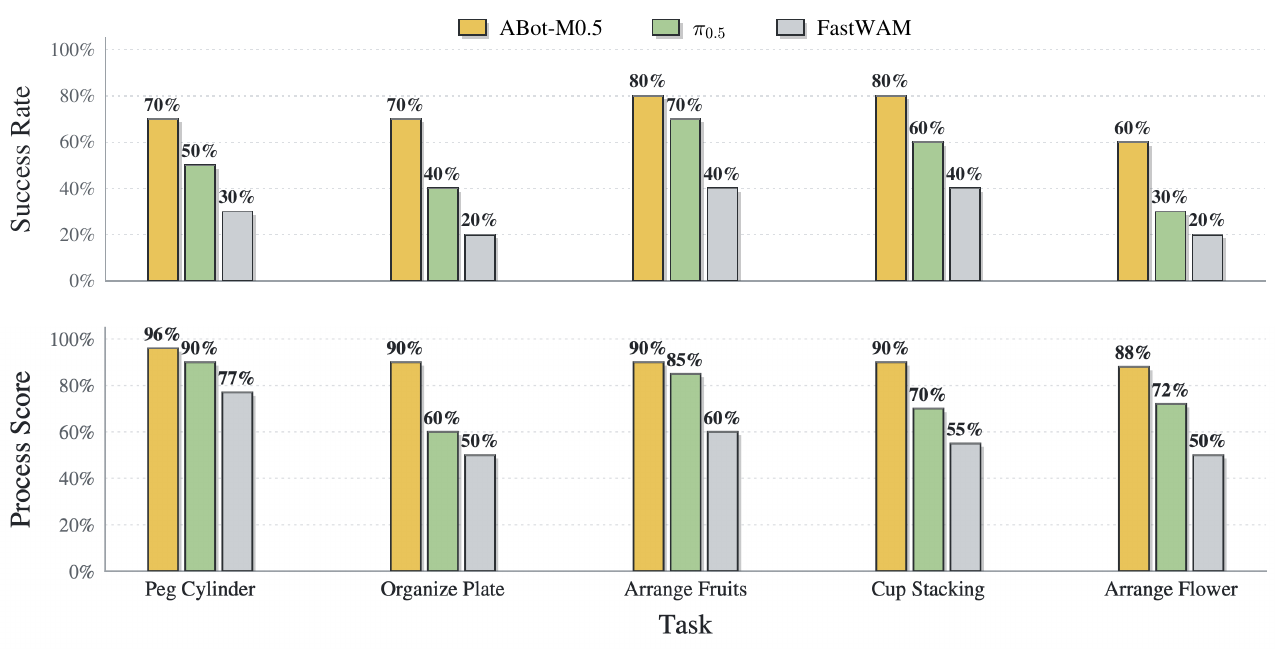}
    \caption{\textbf{Results of real-world experiments.} The performance of \ABotWAM{} was evaluated on five manipulation tasks from two task categories. Our method achieved state-of-the-art performance across all metrics.}
    \label{fig:real_world_result}
\end{figure}

\begin{figure}[!h]
    \centering
    \includegraphics[width=1.0\linewidth]{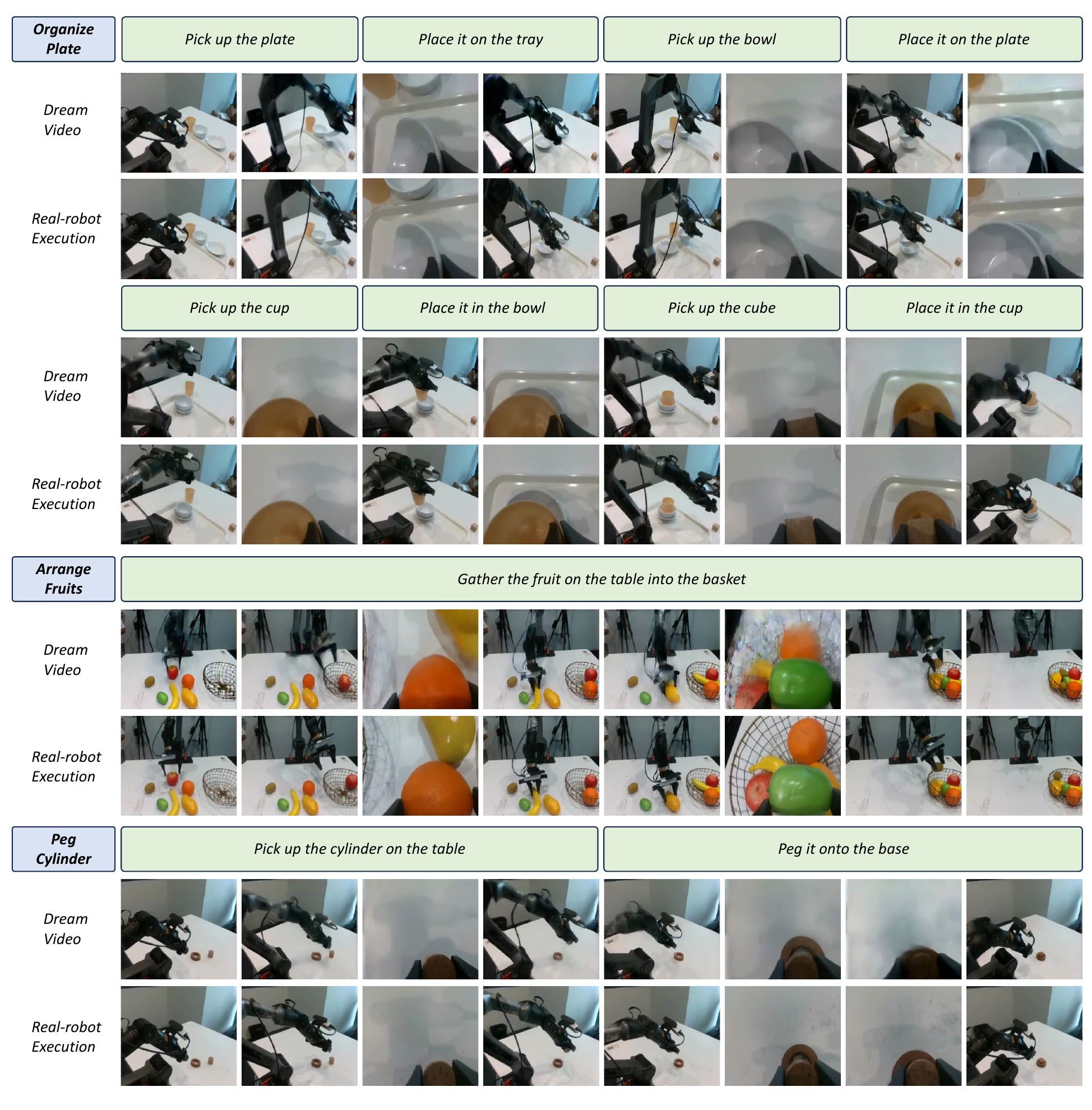}
    \caption{\textbf{Visualization of some real-world results of \ABotWAM{}.} We show video frames of both real camera observations and model-dreamed scenarios for each sample.}
    \label{fig:real_world_demonstration}
\end{figure}
\subsection{Real-World Experiments}
\label{subsec:real_world}

Finally, we evaluated \ABotWAM{} on a real robotic platform to assess its transferability from simulation to physical deployment. The evaluation focused on two complementary capabilities: \textbf{fine-grained manipulation} requiring high spatial precision, and \textbf{long-horizon embodied control} requiring sustained reasoning and execution over multiple interaction steps.

\paragraph{Experimental Setup}
Experiments were conducted on an Agilex Piper robotic arm (6-DoF, single arm). To evaluate data efficiency under practical deployment conditions, only \textbf{50 real-world demonstrations} were collected for each task. We compare our method with $\pi_{0.5}$~\cite{intelligence2025pi05visionlanguageactionmodelopenworld} and Fast-WAM~\cite{yuan2026fastwamworldactionmodels} on two kinds of tasks:

(1) Fine manipulation task:
We evaluated the model on a peg cylinder task, where the robot was instructed to insert a cylinder block into its corresponding hole. Successful execution requires accurate instruction grounding, precise visual localization, and stable closed-loop control, making the task particularly sensitive to errors in vision--action alignment.

(2) Long-horizon tasks:
We further evaluated \ABotWAM{} on three multi-stage manipulation tasks: plate arrangement, fruit arrangement, and flower arrangement. These tasks require the robot to interpret language instructions, continuously reason over evolving scene states, and coordinate a sequence of object interactions to accomplish the final goal.

\paragraph{Experimental Results}
As shown in~\Cref{fig:real_world_result}, \ABotWAM{} consistently achieved strong performance across all evaluated tasks. Taking the fine-grained Peg Cylinder task as an example, our model achieved a 70\% success rate and a 96\% process score, which are clearly higher than those of $\pi_{0.5}$ (50\% and 90\%) and FastWAM (30\% and 77\%). As the last case shown in ~\Cref{fig:real_world_demonstration}, the model reliably localized the target object and completed precise insertion under language guidance. This result suggests that the proposed latent action representation effectively makes up for the limited ability of coarse-grained video representations to capture fine motion dynamics, leading to much better precision in manipulation.

The long-horizon tasks present a different challenge, requiring the robot to maintain task context while continuously adapting to environmental changes. Across the Organize Plate, Arrange Fruits, Cup Stacking, and Arrange Flower tasks, \ABotWAM{} executed coherent multi-step behaviors. Specifically, it achieved success rates of 70\%, 80\%, 80\%, and 60\% respectively, while FastWAM only reached 20\% to 40\% in these complex scenarios. Moreover, our model maintained high process scores (above 88\%) across all these tasks. This indicates that the WAM architecture effectively models long-range temporal dependencies and supports stable sequential decision making, avoiding the build-up of errors often seen in baseline models.

Notably, the same model performs well on both fine-grained and long-horizon manipulation without task-specific changes. This flexibility highlights the benefit of Dual-level MoT for separating different modalities and action distributions. Furthermore, the consistently high process scores demonstrate that Dream-Forcing successfully reduces the gap between training and inference, resulting in more robust execution after deployment. Overall, these results provide strong evidence that \ABotWAM{} can be effectively transferred to real-world robotic deployment with consistent and reliable performance.

\section{Conclusion and Future Work}
\label{sec:conclusion}

We present ABot-M0.5, a unified World Action Model that jointly optimizes temporal alignment and action abstraction to build robust mobile manipulation agents. 
To address the structural mismatches in existing WAMs, we introduce three core innovations. First, we introduce intermediate latent actions to bridge the granularity gap between video and actions. Second, we design a dual-level Mixture-of-Transformers architecture to separate different action spaces. Finally, we propose Dream Forcing to bridge the critical train-test gap in action prediction by conditioning action learning on self-dreamed latents.
Experiments across RoboCasa365, RoboTwin 2.0, LIBERO/LIBERO-Plus, and real-world tasks show that ABot-M0.5 achieves state-of-the-art performance in both long-horizon mobile and fine-grained manipulation, surpassing leading VLA and WAM baselines. These results confirm that extending world models to mobile manipulation is achievable not merely by scaling, but through the systematic engineering of future prediction, action decoupling, and deployment-time robustness.

Looking forward, we aim to scale this framework to broader real-world data and diverse robot embodiments, moving beyond controlled lab settings to validate its generalization in complex, unstructured real-world mobile manipulation. We also plan to systematically investigate the Scaling Laws of WAMs, exploring how task performance scales with model capacity, data volume, and compute budget, thereby providing theoretical guidance for future architecture design. To tackle open-ended long-horizon tasks, we will design more efficient memory mechanisms capable of compressing and retrieving long-term context without catastrophic forgetting. Furthermore, we will focus on optimizing inference time through advanced decoding strategies and hardware-aware acceleration, ensuring real-time policy execution on edge devices. Together, these efforts will push the boundaries of embodied intelligence, transitioning WAMs from simulation and lab prototypes to truly deployable, general-purpose physical agents.

\clearpage
\section{Contributions}
\label{sec:contributions}
\setlength{\parskip}{0pt} 
\setlength{\itemsep}{0pt} 
\setlength{\parsep}{0pt}  
\raggedcolumns

Author contributions in the following areas are as follows:
\begin{itemize}
    \item \textbf{Data Collection \& Standardization:} Yandan Yang, Ronghan Chen, Yuzhi Chen, Haoyun Liu, Dekang Qi
    \item \textbf{Model \& Training:} Ronghan Chen, Zuojin Tang, Tong Lin, Yandan Yang
    \item \textbf{Post-Training \& Evaluation:} Zuojin Tang, Tianlun Li, Haoning Wu, Ronghan Chen, Tong Lin, Mingxin Wang, Bin Hu
    \item \textbf{Real-Robot Experiments \& Deployment:} Dongjie Huo, Lulu Zheng, Botai Yuan
    \item \textbf{Writing:} Yandan Yang, Ronghan Chen, Zuojin Tang, Dekang Qi, Haoyun Liu
    \item \textbf{Challenge Submission:} Yanqing Zhu$^\ddagger$, Wei Mei, Yuze Xuan, Haolong Yang, Dongjie Huo
    \item \textbf{Project Lead:} Xinyuan Chang
    \item \textbf{Advisor:}  Mu Xu$^\dagger$, Zhiheng Ma
\end{itemize}

\vspace{10pt}

\paragraph{Acknowledgments}
We thank Jian Zhang, Ziqiao Li, and Chunlong Lv for their resource support. We also thank Zheng Wu, Zheng Zhang, Dazhi Zhang, Zhiming Sun, and Hongyu Pan for their contributions to data collection and processing. Furthermore, we deeply appreciate the hardware support provided by Xuan Zhou and Yufeng Wang.

{\renewcommand{\thefootnote}{\fnsymbol{footnote}}
\footnotetext[2]{Corresponding Author: xumu.xm@alibaba-inc.com}\footnotetext[3]{Lead for Challenge Submission.}}

\clearpage

\bibliographystyle{plainnat}
\bibliography{main}

\clearpage
\beginappendix

\section{Appendix}

\let\clearpage\relax

\end{document}